\documentclass[10pt,twocolumn,letterpaper]{article}

\usepackage{iccv}
\usepackage{times}
\usepackage{epsfig}
\usepackage{graphicx}
\usepackage{amsmath}
\usepackage{amssymb}

\usepackage{url}
\usepackage{multirow}
\usepackage{framed}
\definecolor{shadecolor}{rgb}{0.9,0.9,0.9}

\usepackage[pagebackref=true,breaklinks=true,letterpaper=true,colorlinks,bookmarks=false]{hyperref}

\newcommand{\bx} {{\bf x }}

\newcommand{\bh} {{\bf h }}

\newcommand{\by} {{\bf y }}

\iccvfinalcopy % *** Uncomment this line for the final submission

\def\iccvPaperID{893} % *** Enter the CVPR Paper ID here

\ificcvfinal\fi
\begin{document}

%%%%%%%%% TITLE
\title{Deep Learning Face Attributes in the Wild\thanks{This work has been accepted to appear in ICCV 2015. This is the pre-printed version. Content may slightly change prior to the final publication.}}

\author{Ziwei Liu$^1$ \hspace{10pt} Ping Luo$^1$ \hspace{10pt} Xiaogang Wang$^2$ \hspace{10pt} Xiaoou Tang$^1$\\
$^1$Department of Information Engineering, The Chinese University of Hong Kong \\
$^2$Department of Electronic Engineering, The Chinese University of Hong Kong \\
{\tt\small \{lz013,pluo,xtang\}@ie.cuhk.edu.hk, xgwang@ee.cuhk.edu.hk}
% \and
}

\maketitle
\thispagestyle{empty}

%%%%%%%%% ABSTRACT
\begin{abstract}

   Predicting face attributes in the wild is challenging due to complex face variations. We propose a novel deep learning framework for attribute prediction in the wild. It cascades two CNNs, LNet and ANet, which are fine-tuned jointly with attribute tags, but pre-trained differently. LNet is pre-trained by massive general object categories for face localization, while ANet is pre-trained by massive face identities for attribute prediction. This framework not only outperforms the state-of-the-art with a large margin, but also reveals valuable facts on learning face representation.

   (1) It shows how the performances of face localization (LNet) and attribute prediction (ANet) can be improved by different pre-training strategies.
   (2) It reveals that although the filters of LNet are fine-tuned only with image-level attribute tags, their response maps over entire images have strong indication of face locations. This fact enables training LNet for face localization with only image-level annotations, but without face bounding boxes or landmarks, which are required by all attribute recognition works.
	 %With a novel fast feed-forward algorithm, the proposed framework can recognize attributes in images with arbitrary sizes in real time.
   %
   (3) It also demonstrates that the high-level hidden neurons of ANet automatically discover semantic concepts after pre-training with massive face identities, and such concepts are significantly enriched after fine-tuning with attribute tags. Each attribute can be well explained with a sparse linear combination of these concepts.
   %By analyzing such combinations, attributes show clear grouping patterns, which could be well interpreted semantically.

\end{abstract}

\vspace{-20pt}

%%%%%%%%% BODY TEXT
%------------------------------------------------------------------------
\section{Introduction}

\begin{figure}[t]
  \centering
  \includegraphics[width=0.45\textwidth]{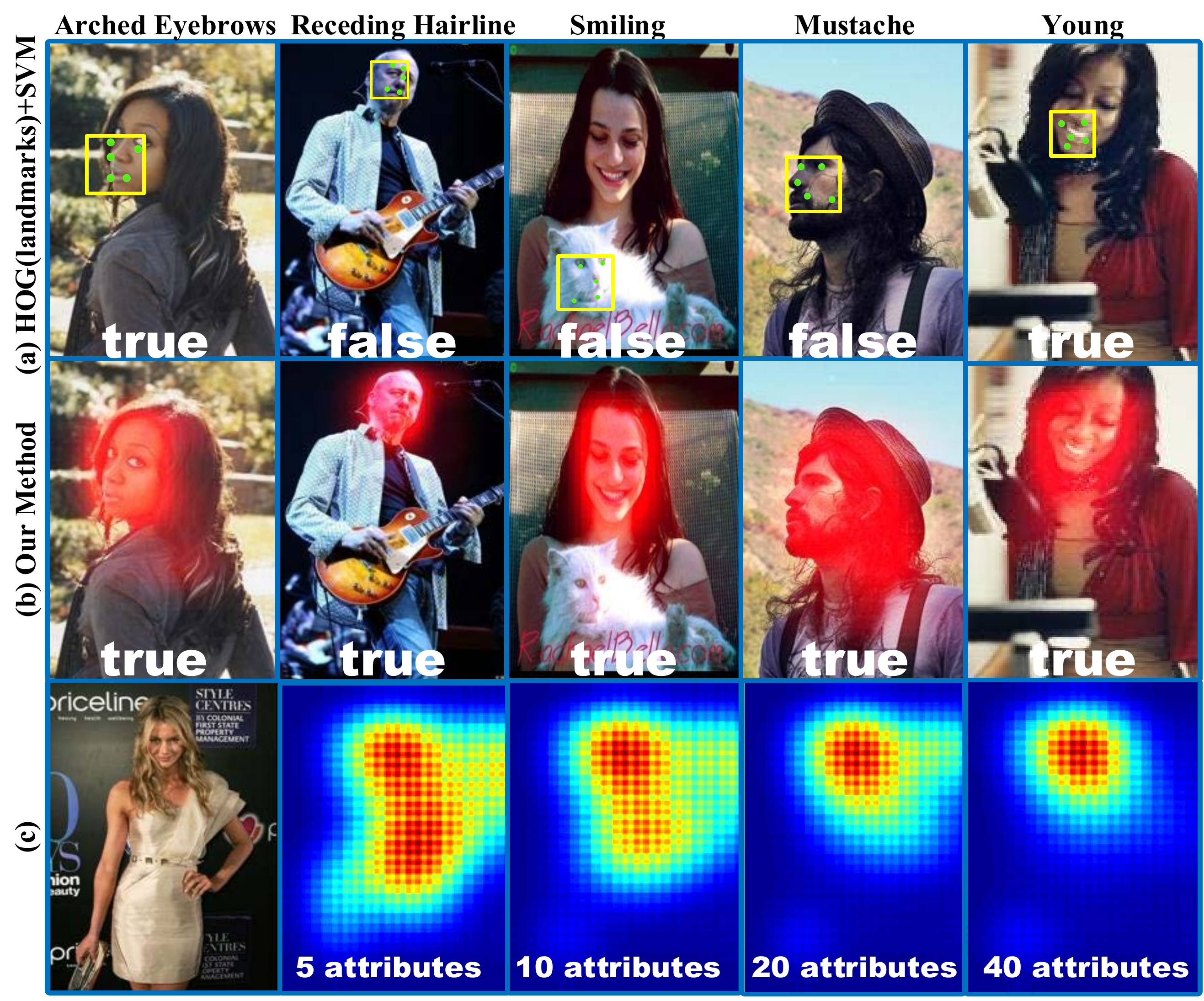}
  \caption{\footnotesize (a) Inaccurate localization and alignment lead to prediction errors on attributes by existing methods (b) LNet localizes face regions by averaging the response maps of attribute filters. ANet predicts attributes without alignment (c) Face localization with the averaged response map when LNet is trained with different numbers of attributes. \textbf{(Best viewed in color)}}
  \vspace{-15pt}
  \label{fig:intro}
\end{figure}

Face attributes are beneficial for multiple applications such as face verification \cite{kumar2009attribute, berg2013poof, song2014exploiting}, identification \cite{manyam2011two}, and retrieval. Predicting face attributes from images in the wild is challenging, because of complex face variations such as poses, lightings, and occlusions as shown in Fig.\ref{fig:intro}.

Attribute recognition methods are generally categorized into two groups: global and local methods.
Global methods extract features from the entire object, where accurate locations of object parts or landmarks are not required. They are not robust to deformations of objects \cite{razavian2014cnn}. Recent local models \cite{kumar2009attribute, bourdev2011describing, chung2012deep, berg2013poof, luo2013deep, zhang2013panda} first detect object parts and extract features from each part. These local features are concatenated to train classifiers. For example, Kumar \etal \cite{kumar2009attribute} predicted face attributes by extracting hand-crafted features from ten face parts. Zhang \etal \cite{zhang2013panda} recognized human attributes by employing hundreds of poselets \cite{bourdev2011describing} to align human body parts.
These local methods may fail when unconstrained face images with complex variations are present, which makes face localization and alignment difficult.
As shown in Fig.\ref{fig:intro} (a), HOG+SVM fails because the faces or landmarks are wrongly localized or misaligned. Thus the features are extracted at wrong positions \cite{sun2013deep}. Recent research shows that face localization and alignment are still not well solved problems, especially in the wild condition, although much progress has been achieved in the past decade. It is also proved by our experimental result.

%It heavily depends on the precision of face and landmark detections, which are not reliable in web images. Fig.\ref{fig:intro} (a) shows the results of state-of-the-art face detection \cite{li2013learning} and alignment \cite{sun2013deep} and the attribute predictions of HOG (landmark)+SVM on challenging images. Most of them fail because features are extracted at wrong landmark positions.
%
%Although this pipeline is suitable for controlled environment, it has drawbacks when dealing with web images. It heavily depends on the precision of face and landmark detections, which are not reliable in web images. Fig.\ref{fig:intro} (a) shows the results of state-of-the-art face detection \cite{li2013learning} and alignment \cite{sun2013deep} and the attribute predictions of HOG (landmark)+SVM on challenging images. Most of them fail because features are extracted at wrong landmark positions.
%
%Face detection also has ambiguity.
%In the third image of Fig.\ref{fig:intro} (a), the face detector confuses the cat face and the human face, as they appear similarly in the HOG space.

%This work addresses face attributes in the wild by revisiting global method and proposes a novel deep learning framework, which has novelties in three aspects.
%

This work revisits global methods by proposing a novel deep learning framework, which integrates two CNNs, LNet and ANet, where LNet locates the \emph{entire face region} and ANet extracts high-level face representation from the located region.
%The pre-training and fine-tuning of both LNet and ANet are carefully designed, making
%
%, and its novelties are in three aspects.
%
The novelties are in three aspects.
Firstly, LNet is trained in a \emph{weakly supervised manner, \ie only image-level attribute tags of training images are provided}, making data preparation much easier. This is different from training face and landmark detectors, where face bounding boxes and landmark positions are required.
%In this case, the annotations in the steps of object location and attribute classification are the same.
%
%Different from training face detectors with positive (face) and negative (non-face) samples,
LNet is pre-trained by classifying \emph{massive general object categories}, such that its pre-trained features have good generalization capability on handling large background clutters. LNet is then fine-tuned by attributes tags.
We demonstrate that features learned in this way are effective for face localization and also can distinguish subtle differences between human faces and analogous patterns, such as a cat face.

%and ANet are first pre-trained differently and then jointly trained with attribute labels.

%Firstly, it does not rely on face and landmark detection. Instead, it cascades two CNNs, LNet and ANet, where LNet locates the entire face region and ANet extracts face representation from the located face region.

%in which one (LNet) to locate face region and the other (ANet) to extract high-level face representation from the entire located face region (without landmarks) for attribute prediction.  \textit{Training LNet and ANet is in a weakly supervised manner, i.e.  only attribute tags of training images are provided}. This is fundamentally different from training face and landmark detectors, where face bounding boxes and landmark positions are needed.  It makes the preparation of training data much easier. LNet and ANet are first pre-trained differently and then jointly trained with attribute labels.

Secondly, ANet extracts discriminative face representation, making attribute recognition from the entire face region possible. ANet is pre-trained by classifying \emph{massive face identities} and is fine-tuned by attributes. We show that the pre-training step enables ANet to account for complex variations in the unconstrained face images.

% is pre-trained by classifying massive face identities
%
%different pre-training and fine-tuning strategies are designed for  LNet and ANet. Different from training face detectors with positive (face) and negative (non-face) samples, LNet is pre-trained by classifying massive general object categories. Thus, its pre-trained features have good generalization capability on handling various background clutters. LNet is then fine-tuned by predicting attributes. Features learned by attribute prediction can capture rich face variations and are effective for face localization. It also can better distinguish subtle differences between human faces and analogous patterns, such as a cat face. ANet is pre-trained by classifying massive face identities, to obtain discriminative face representation. Then it is fine-tuned by the attribute prediction task.

Thirdly, within the rough locations of face regions provided by LNet,
averaging the predictions of multiple patches can improve the performance.
A simple way is to evaluate the feed-forward pass for each single patch.
%
%patch-by-patch scanning and max-pooling with ANet are required.
However, it is slow and has a lot of redundant computation.  A novel fast feed-forward scheme is proposed to replace patch-by-patch evaluation.
It evaluates images with arbitrary sizes with only one-pass feed-forward operation.
%If filters are globally shared, this can be done by convolving images with filters.
It becomes non-trivial if the filters are \emph{locally shared}, while studies \cite{taigman2014deepface, sun2014deep} showed that locally shared filters perform better in face related tasks. This is solved by proposing an interweaved operation.

Besides proposing new methods, our framework also reveals \textbf{valuable facts} on learning face representation. They not only motivate this work but also benefit future research on face and deep learning.
(1) It shows how pre-training with \emph{massive object categories} and \emph{massive identities} can improve feature learning for face localization and attribute recognition, respectively.
(2) It demonstrates that although filters of LNet are fine-tuned by attribute tags, their response maps over the entire image have strong indication of face location.
Good features for face localization should be able to capture rich face variations, and \emph{more supervised information on these variations improves the learning process}.
%
%The example in Fig. \ref{fig:intro} (a) and the experimental result in Fig. \ref{fig:lnet} (d) show that as the number of attributes decreases, the localization capability of learned neurons gets reduced dramatically.
The examples in Fig. \ref{fig:intro} (a) show that as the number of attributes decreases, the localization capability of learned neurons gets reduced dramatically.
(3) ANet is pre-trained with \emph{massive face identities}. It discloses that the pre-trained high-level hidden neurons of ANet implicitly learn and discover sematic concepts that are related to identity, such as race, gender, and age.
%These concepts are significantly expanded after fine-tuning with attributes.
%
It indicates that when a deep model is pre-trained for face recognition, it implicitly learns attributes. The performance of attribute prediction drops without this pre-training stage.
%Each face attribute is well explained by a sparse linear combination of these sematic concepts.
%
% By analyzing the coefficients of such combinations, attributes show clear grouping patterns, which could be well interpreted semantically.

The main \textbf{contributions} are summarized as follows. (1) We propose a novel deep learning framework, which combines \emph{massive objects} and \emph{massive identities} to pre-train two CNNs for face localization and attribute prediction, respectively.
%. Two cascaded CNNs are trained in a weakly supervised manner, which makes it easier to prepare training data from web images.
It achieves state-of-the-art attribute classification results on both the challenging CelebFaces \cite{sun2014deep} and LFW \cite{LFWTech} datasets, improving existing methods by $8$ and $13$ percent, respectively.
(2) A novel fast feed-forward algorithm for CNN with \emph{locally shared filters} is devised.
(3) Our study reveals multiple valuable facts on leaning face representation by deep models.
(4) We also contribute a large facial attribute database with more than eight million attribute labels and it is $20$ times larger than the largest publicly available dataset.

\begin{figure*}[t]
  \centering
  \includegraphics[width=0.9\textwidth]{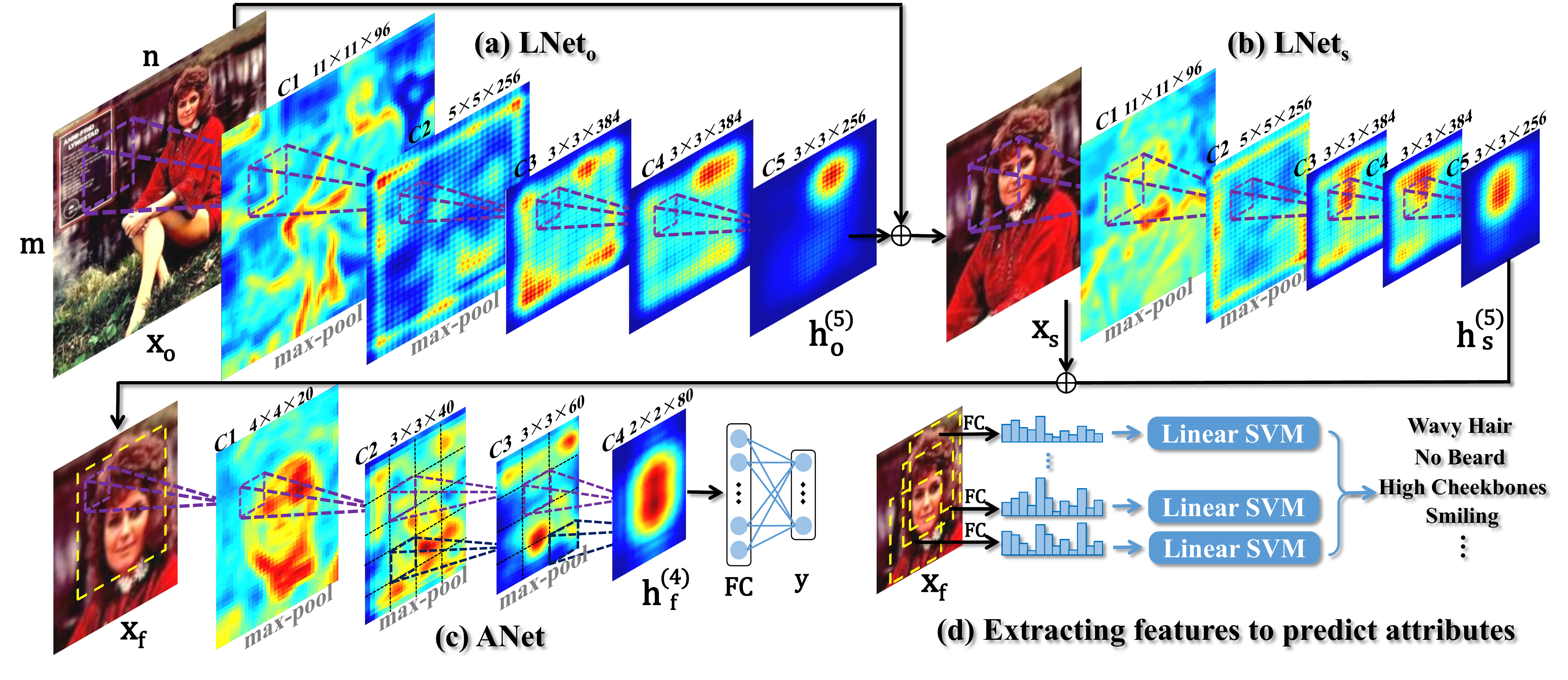}
  \caption{\footnotesize The proposed pipeline of attribute prediction \textbf{(Best viewed in color)}}
  \vspace{-10pt}
  \label{fig:pipeline}
\end{figure*}

%------------------------------------------------------------------------
\subsection{Related Work}

%\textbf{Attribute Prediction with Hand-crafted Features}
Extracting hand-crafted features at pre-defined landmarks has become a standard step in attribute recognition \cite{farhadi2009describing, kumar2009attribute, bourdev2011describing, berg2013poof}.
%
% Farhadi \etal \cite{farhadi2009describing} combined HOG and color histogram to train logistic regression for object search and tagging based on attributes.
%
Kumar \etal \cite{kumar2009attribute} extracted HOG-like features on various face regions to tackle attribute classification and face verification.
To improve the discriminativeness of hand-crafted features given a specific task, Bourdev \etal \cite{bourdev2011describing} built a three-level SVM system to extract higher-level information. %Berg \etal \cite{berg2013poof} combined various hand-crafted features to obtain an intermediate representation for a particular domain.
%
%\textbf{Attribute Prediction with Deep Models}
Deep learning \cite{razavian2014cnn, donahue2013decaf, luo2013deep, zhang2013panda, zhang2014part, krizhevsky2012imagenet, zhou2014object, oquab2015object, bergamo2014self} recently achieved great success in attribute prediction, due to their ability to learn compact and discriminative features.
Razavian \etal \cite{razavian2014cnn} and Donahue \etal \cite{donahue2013decaf} demonstrated that off-the-shelf features learned by CNN of ImageNet \cite{krizhevsky2012imagenet} can be effectively adapted to attribute classification.
Zhang \etal \cite{zhang2013panda} showed that better performance can be achieved by ensembling learned features of multiple pose-normalized CNNs.
%
% Specific network structures have also been designed for attribute prediction.
% Luo \etal \cite{luo2013deep} introduced a deep sum-product architecture to account for occlusion during attribute inference.
%
The main drawback of these methods is that they rely on accurate landmark detection and pose estimation in both training and testing steps.
Even though a recent work \cite{zhang2014part} can perform automatic part localization during test, it still requires landmark annotations of the training data.

%------------------------------------------------------------------------
\section{Our Approach}

\textbf{Framework Overview} Fig.\ref{fig:pipeline} illustrates our pipeline where LNet locates the entire face region in a coarse-to-fine manner as shown in (a) and (b), while ANet extracts features for attribute recognition as shown in (c).

Different from existing works that rely on accurate face and landmark annotations, LNet is trained in a weakly supervised manner with only image-level annotations. Specifically, it is pre-trained with one thousand object categories of ImageNet \cite{deng2009imagenet} and fine-tuned by image-level attribute tags. The former step accounts for background clutters, while the latter step learns features robust to complex face variations. Learning LNet in this way not only significantly reduces data labeling, but also improves the accuracy of face localization.
Both LNet$_o$ and LNet$_s$ have network structures similar to AlexNet \cite{krizhevsky2012imagenet}, whose hyper parameters are specified in Fig.\ref{fig:pipeline} (a) and (b) respectively. The fifth convolutional layer (C5) of LNet$_o$ indicates head-shoulders while C5 of LNet$_s$ indicates faces, with their highly responsed regions in their averaged response maps. Moreover, the input $\bx_o$ of LNet$_o$ is a $m\times n$ image, while the input $\bx_s$ of LNet$_s$ is the head-shoulder region, which is localized by LNet$_o$ and resized to $227\times227$.

%detects the region of head-shoulder and LNet$_s$ detects face region. Both of them have network structures similar to AlexNet \cite{}. Their hyper parameters are specified in Fig.\ref{fig:pipeline} (a) and (b), respectively,

As illustrated in Fig.\ref{fig:pipeline} (c), ANet is learned to predict attributes $\by$ by providing the input face region $\bx_f$,  which is detected by LNet$_s$
and properly resized.
% and resized to $55\times 47$.
Specifically, multi-view versions \cite{krizhevsky2012imagenet} of $\bx_f$ are utilized to train ANet.
Furthermore, ANet contains four convolutional layers, where the filters of C1 and C2 are globally shared and the filters of C3 and C4 are locally shared. The effectiveness of local filters have been demonstrated in many face related tasks \cite{sun2013deep, taigman2014deepface}.
To handle complex face variations, ANet is pre-trained by distinguishing massive face identities, which facilitates the learning of discriminative features.

Fig.\ref{fig:pipeline} (d) outlines the procedure of attribute recognition.
ANet extracts a set of feature vectors (FCs) by cropping overlapping patches on $\bx_f$.
% without rescaling.
An efficient feed-forward algorithm is developed to reduce redundant computation in the feature extraction stage.
SVMs \cite{REF08a} are trained to predict attribute values given each FC.
% Second, SVM \cite{REF08a} predicts attribute values given each feature vector FC.
The final prediction is obtained by averaging all these values, to cope with small misalignment of face localization.

\subsection{Face Localization}

The cascade of LNet$_o$ and LNet$_s$ accurately localizes face regions by being trained on image-level attribute tags. %We will briefly introduce pre-training and fine-tuning, and then discuss face localization in details.

\begin{figure}[t]
  \centering
  \includegraphics[width=0.45\textwidth]{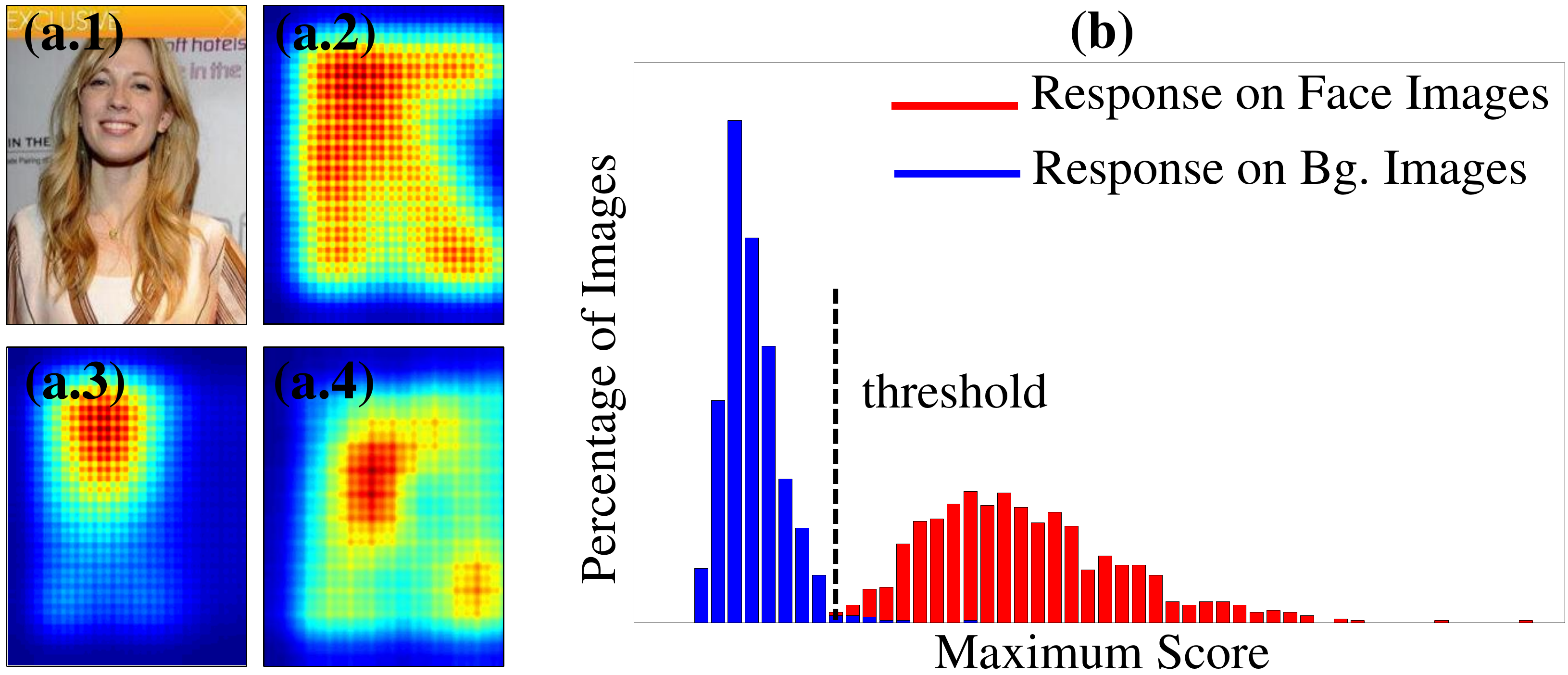}
  \caption{\footnotesize (a.1) Original image. (a.2)-(a.4) are averaged response maps in C5 of LNet$_o$ after pre-training (a.2), fine-tuning (a.3) and directly training from scratch with attribute tags but without pre-training (a.4). (b) Determine threshold.}
  \vspace{-15pt}
  \label{fig:singleface}
\end{figure}

\textbf{Pre-training LNet} Both LNet$_o$ and LNet$_s$ are pre-trained with $1,000$ general object categories
from the ImageNet Large Scale Visual Recognition Challenge (ILSVRC) 2012 \cite{deng2009imagenet}, containing $1.2$ million training images and $50$ thousands validation images.
All the data is employed for pre-training except one third of the validation data for choosing hyper-parameters \cite{krizhevsky2012imagenet}. We augment data by cropping ten patches from each image, including one patch at the center and four at the corners, and their horizontal flips. We adopt softmax for object classification, which is optimized by stochastic gradient descent (SGD) with back-propagation (BP) \cite{le1990handwritten}. As shown in Fig.\ref{fig:singleface} (a.2), the averaged response map in C5 of LNet$_o$ already indicates locations of objects including human faces after pre-training.

\textbf{Fine-tuning LNet} 
Both LNet$_o$ and LNet$_s$ are fine-tuned with attribute tags.
Additional output layers are added to the LNets individually for fine-tuning and then removed for evaluation.
LNet$_o$ adopts the full image $\bx_o$ as input while LNet$_s$ uses the highly responsed region $\bx_s$ in the averaged response map in C5 of LNet$_o$ as input, which roughly respond to head-shoulders.
%
%Similar to the pre-training of LNet+, we add two fully-connected layers to both LNet$_o$ and LNet$_s$, where the weight matrixes are initialized randomly.
%
The cross-entropy loss is used for attribute classification, \ie $L=\sum_{i=1}\by_i\log p(\by_i|\bx) + (1 - \by_i)\log\big(1 - p(\by_i|\bx)\big)$, where $p(\by_i=1|\bx)=\frac{1}{1+\exp (-f(\bx))}$ is the probability of the $i$-th attribute given image $\bx$.
%It is also optimized by SGD with BP.
As shown in Fig.\ref{fig:singleface} (a.3), the response maps after fine-tuning become much more clean and smooth, indicating that the filters learned by attribute tags can detect face patterns with complex variations. To appreciate the effectiveness of pre-training, we also include the averaged response map in C5 of being directly trained from scratch with attribute tags but without pre-training in Fig.\ref{fig:singleface} (a.4).
It cannot separate face regions from background and other body parts well.

\textbf{Thresholding and Proposing Windows} %We have demonstrated in Fig.\ref{fig:singleface} (a) that high responses of C5 in LNet indicate face regions.
We show that the responses of C5 in LNet are discriminative enough to separate faces and background by simply searching a threshold, such that a window with response larger than this threshold corresponding to face and otherwise is background.
To determine the threshold, we select $2000$ images, each of which contains a single face, and $2000$ background images from SUN dataset \cite{xiao2010sun}. For each image, EdgeBox \cite{zitnick2014edge} is adopted to propose $500$ candidate windows, each of which is measured by a score that sums over its response values normalized by its window size. A larger score indicates the localized pattern is more likely to be a face. Each image is then represented by the maximum score over all its windows. In Fig.\ref{fig:singleface} (b), the histogram of the maximum scores shows that these scores clearly separate face images from background images. The threshold is chosen as the decision boundary as shown in Fig.\ref{fig:singleface} (b). More results are given in Fig.\ref{fig:localizationmap} (a), showing that the above strategy can precisely localize face within a single test image. Since each training image only contains one single face, we localize a face region using the window with the largest score during training.

\textbf{Pruning Multiple Faces within a single Window}. For some challenging cases in the testing stage, it encounters difficulty when multiple faces are presented within a single window, such that there may be multiple regions with high responses. We predict attributes based one face region which generates the largest response\footnote{In CelebFaces and LFW, it is assumed that each image has a ``dominant'' face, based on which the attribute tags were labeled by users.}.
Similar to \cite{rodriguez2014clustering}, a fast density peak identification technique is devised.
It calculates a special geodesic distance for each position $i$ in the response map,
$d_{i} = (\rho_{i}^{2} + \sigma_{i}^{2})^{1/2}$, where $\rho_{i}$ is the density intensity in position $i$, $\sigma_{i} = \min_{j:\rho_{j}>\rho_{i}}(s_{ij})$ and $s_{ij}$ is the spatial distance between $i$ and $j$.
% $\sigma_{i}$ measures its distance to the nearest position which has a larger density intensity.
%
Then density peaks are identified by selecting extreme large $d_{i}$.
This process can be further accelerated, as the averaged response map in C5 is sparse.
We propose the correct window by cropping the region with the highest density.

\begin{figure}[t]
  \centering
  \includegraphics[width=0.45\textwidth]{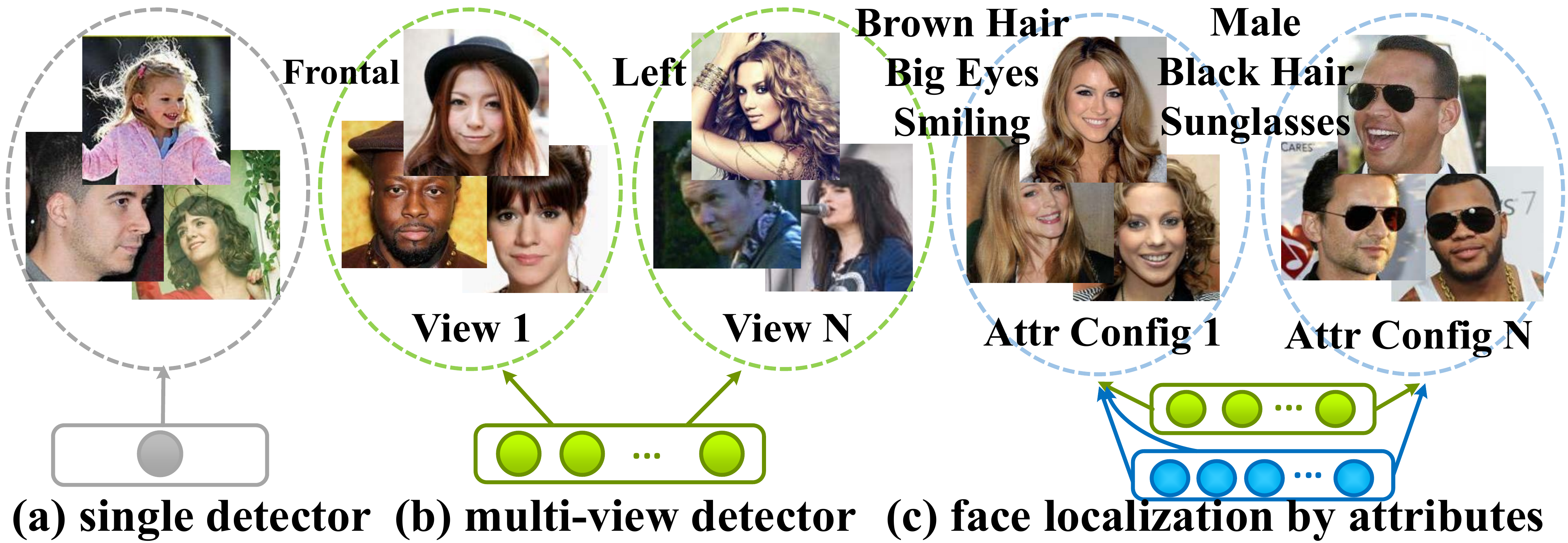}
  \caption{\footnotesize Face localization by attributes}
  \vspace{-15pt}
  \label{fig:detector}
\end{figure}

To understand why rich attribute information enables accurate face localization, one could consider the examples in Fig.\ref{fig:detector}. If only a single detector \cite{li2013learning, mathias2014face} is used to classify all the positive and negative samples in Fig.\ref{fig:detector} (a), it is difficult to handle complex face variations.
Therefore, multi-view face detectors \cite{yang2014aggregate} were developed in Fig.\ref{fig:detector} (b), \ie face images in different views are handled by different detectors. View labels were used in training detectors and the whole training set is divided into subsets according to views. If views are treated as one type of face attributes, learning face representation by predicting attributes with deep models actually extends this idea to extreme. As shown in Fig.\ref{fig:detector} (c), a filter (or a group of filters) functions as a detector of an attribute. When a subset of neurons are activated, they indicate the existence of face images with a particular attribute configuration. The neurons at different layers can form many activation patterns, implying that the whole set of face images can be divided into many subsets
based on attribute configurations, and each activation pattern corresponds to one subset (\eg `pointy nose', `rosy cheek', and `smiling'). Therefore, it is not surprising that filters learned by attributes lead to effective representations for face localization.
% By simply averaging and thresholding response maps, good face localization is achieved.

\subsection{Attribute Prediction}

As shown in Fig.\ref{fig:pipeline} (c) and (d), ANet is learned to extract features and SVM classifiers are used to predict attributes.
Specifically, in the pre-training stage, ANet is trained by classifying massive face identities.
In the fine-tuning stage, we first extend the localized face region, which is properly resized, with a small factor to incorporate more context information.
Then, multiple patches are cropped from the enlarged face region and utilized as inputs of ANet.
% In the fine-tuning stage, we first extend the localized face bounding box with a small factor to incorporate more context information.
% Then, multiple patches are cropped from the enlarged face region, each of which is resized to $55\times47$ and utilized as input of ANet.
ANet is fine-tuned by attributes to learn the high-level feature $\mathrm{FC}$.
Furthermore, as shown in Fig.\ref{fig:pipeline} (d), each feature vector is adopted to train SVM classifier for attribute prediction. The above strategy is similar to the multi-view data augmentation \cite{krizhevsky2012imagenet}, increasing the robustness of attribute recognition. In the testing stage, attributes are predicted by averaging the SVM scores over all the patches.

%the high-level feature, $\mathrm{FC}$, we first crop multiple patches from the original localized face region, which is the face region localized by LNet$_s$ without re-scaling to $55\times47$. We also enlarge the region with a small factor to incorporate more context information. Then we extract feature on each region using ANet, as shown in Fig.\ref{fig:pipeline} (d). The above scheme can deal with face misalignment, and thus improve robustness and accuracy of attribute recognition.

\begin{figure*}[t]
  \centering
  \includegraphics[width=0.9\textwidth]{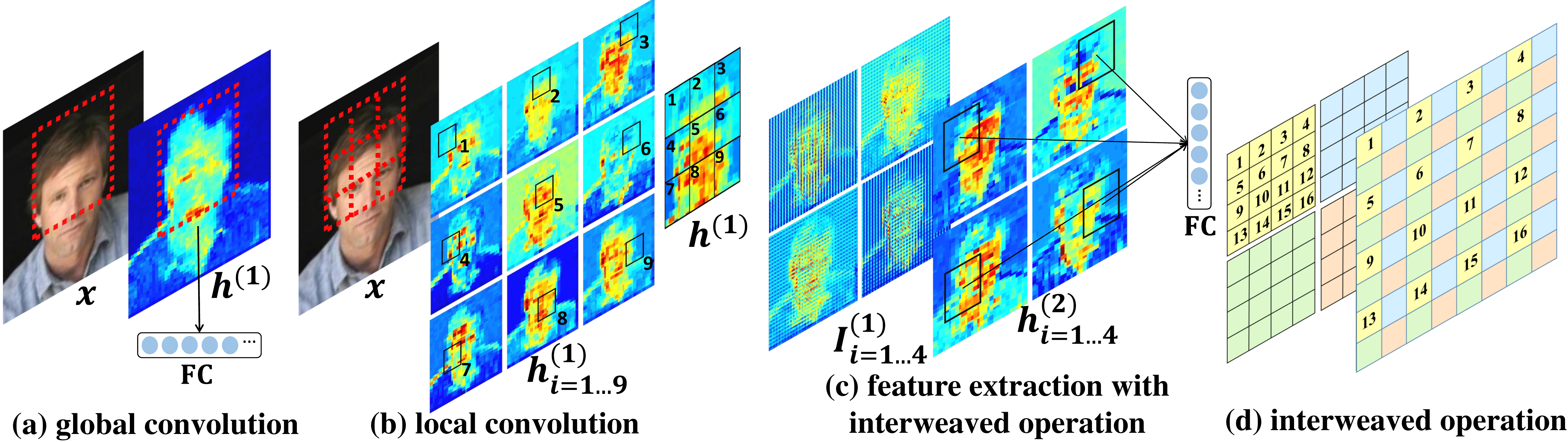}
  \caption{\footnotesize Detailed pipeline of efficient feature extractions in ANet.}
  \label{fig:interweaved}
\end{figure*}

\textbf{Pre-training of ANet} We introduce how to learn discriminative features by pre-training ANet with a large number of identities.
We select eight thousand face identities from the CelebFaces \cite{sun2014deep} dataset, where each identity has around twenty images. There are over $160$ thousand training images in total.
%We follow \cite{sun2014deep} to preprocess and augment these data.
%
A simple way to train ANet is to classify eight thousand categories with the softmax loss.
However, it is challenging because the number of samples of each identity is limited to maintain the intra-class invariance.
To improve intra-class invariance, we employ the similarity loss similar to \cite{sun2014deep, hadsell2006dimensionality}. It decreases the distances between samples of the same identity. We have $L=\sum_{i=1, \by_i=\by_j}^{|D|}\|\mathrm{FC}_i-\mathrm{FC}_j\|_{2}^{2}$, where $\mathrm{FC}_i$ and $\mathrm{FC}_j$ denote the feature vectors of the $i$-th and $j$-th face images respectively, and $\by_i=\by_j$ indicates the identities of these samples are the same.
In summary, ANet is pre-trained by combining the softmax loss and the similarity loss. 
% More details can be found in the \emph{supplementary material}.

\textbf{Efficient Feature Extractions}
In test, ANet is evaluated on multiple patches of the face region as shown in Fig.\ref{fig:pipeline} (d), leading to redundant convolutional
computations because of the large overlaps in these patches.
When all the filters are globally-shared, the computational cost can be reduced by applying \cite{he2014spatial}, which convolves the filters in the input image and then obtains a feature vector for each patch by pooling over the last convolutional layer.
Given a simple example with one convolutional layer as shown in Fig.\ref{fig:interweaved} (a), the feature vector $\mathrm{FC}$ for each patch (\eg rectangle in red) can be extracted by pooling in the corresponding region of the response map $\bh^{(1)}$, without evaluating convolutions in the input image patch-by-patch.
Therefore, it shares the convolutions for every patch.
%It is able to evaluate input image with arbitrary size, when the filters are globally-shared.

However, this scheme is not applicable when we have more than two convolutional layers whose filters are locally-shared. An example is illustrated in Fig.\ref{fig:interweaved} (b), where each patch is equally divided into $3\times 3=9$ cells and we learn different filters for different cells.
To reduce computations in the first convolutional layer, each local filter can be applied on the entire image, resulting in the response map with nine channels, \ie $\bh^{(1)}_{i}$ and $i=1...9$.
The final response map $\bh^{(1)}$ is obtained by cropping and padding the regions (\ie rectangles in black) in these 9 channels.
As a result, each feature vector $\mathrm{FC}$ can be pooled from $\bh^{(1)}$, without convolving the input image patch-by-patch.
Nevertheless, since $\bh^{(1)}$ is corresponded to a patch of the input image, the succeeding local convolutions have to be handled patch-by-patch, leading to redundant computations.

%\textbf{Interweaved Operation}
%
To this end, we propose an \emph{interweaved operation}, which is a fast feed-forward method for CNN with locally-shared filters.
%can account for misalignment without cropping multiple patches.
Suppose we have four local filters in the next locally convolutional layer and each filter is applied on $2\times 2$ cells of $\bh^{(1)}$ as shown in (b). These cells are the receptive fields of the filters, including $\{1,2,4,5\}$, $\{2,3,5,6\}$, $\{4,5,7,8\}$, and $\{5,6,8,9\}$.
Instead of directly applying the local filters on $\bh^{1}$, the interweaved operation generates an interweaved map $I^{(1)}_{i}$ for each filter, where $i=1...4$. 
Each local filter is then apply on its corresponding interweaved map. Since the interweaved map capturing the entire image, each local filter is turned into a global filter such that its computation can be shared across different patches.

Specifically, each interweaved map, \eg $I^{(1)}_{1}$, is achieved by padding the cells of the corresponding channels in an interweaved manner, \eg $\bh^{(1)}_{i=\{1,2,4,5\}}$, as shown in Fig.\ref{fig:interweaved} (d).
All of the interweaved maps are illustrated in Fig.\ref{fig:interweaved} (c). After that, each of the four local filters is applied on its corresponding interweaved map, leading to four response maps $\bh^{(2)}_{i}$, where $i=1...4$.
As a result, the feature vector $\mathrm{FC}$ is pooled and concatenated from the receptive fields of the filters, which are the rectangles in black as shown in (c).

Intuitively, instead of padding cells according to the receptive fields of all the local filters (\eg $\bh^{(1)}$ in (b)), which has to be performed in a patch-by-patch way,
the interweaved operation pads the cells with respect to the receptive field of each local filter over the entire image.
It enables extracting multiple feature vectors with only one-pass of feed-forward evaluation.
This operation can be repeated when more locally convolutional layers are added.
The proposed feature extraction scheme has achieved $6\times$ speedup empirically when compared with patch-by-patch scanning.
It is applicable to CNNs with local filters and compatible to all existing CNN operations.

\section{Experiments}

\begin{figure*}[t]
  \centering
  \includegraphics[width=0.9\textwidth]{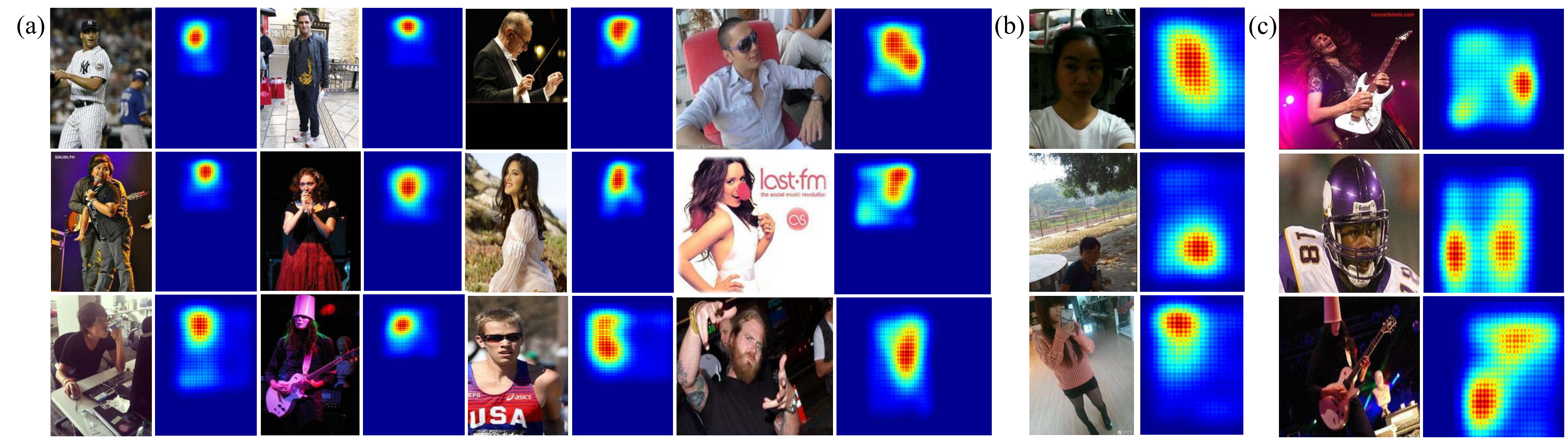}
  \caption{\footnotesize Averaged response maps of LNet, including (a) CelebA, (b) MobileFaces, (c) some failure cases. \textbf{(Best viewed in color)}}
  \vspace{-10pt}
  \label{fig:localizationmap}
\end{figure*}

\textbf{Large-scale Data Collection}
We construct two face attribute datasets, namely CelebA and LFWA, by labeling images selected from two challenging face datasets, CelebFaces \cite{sun2014deep} and LFW \cite{LFWTech}. CelebA contains ten thousand identities, each of which has twenty images. There are two hundred thousand images in total. LFWA has $13,233$ images of $5,749$ identities. Each image in CelebA and LFWA is annotated with forty face attributes and five key points by a professional labeling company. CelebA and LFWA have over eight million and five hundred thousand attribute labels, respectively.

CelebA is partitioned into three parts. Images of the first eight thousand identities (with $160$ thousand images) are used to pre-train and fine-tune ANet and LNet, and the images of another one thousand identities (with twenty thousand images) are employed to train SVM.
%, as introduced in Sec.\ref{sec:learning}.
The images of the remaining one thousand identities (with twenty thousand images) are used for testing.
%Note that LNets and ANet utilize the same subset for fine-tuning.
%
LFWA is partitioned into half for training and half for testing. Specifically, $6,263$ images are adopted to train SVM and the remaining images for test. When being evaluated on LFWA, LNet and ANet are trained on CelebA.

\begin{figure}[t]
  \centering
  \includegraphics[width=0.45\textwidth]{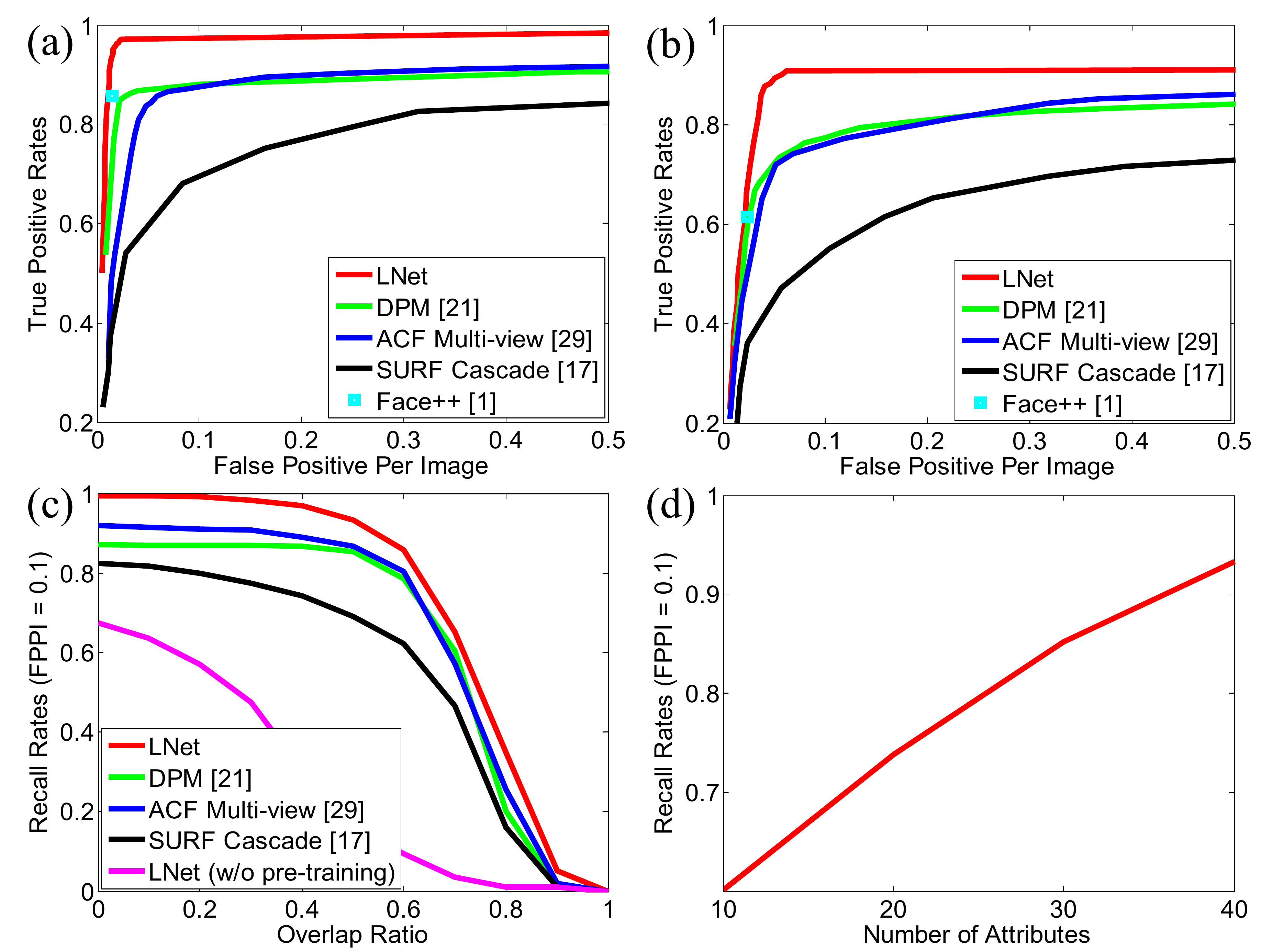}
  \caption{\footnotesize ROC curves on (a) CelebA (b) MobileFaces. (c) Recall rates \wrt overlap ratio ($FPPI = 0.1$). (d) Recall rates \wrt number of attributes ($FPPI = 0.1$)}
  \vspace{-10pt}
  \label{fig:lnet}
\end{figure}

\textbf{Methods for Comparisons}
The proposed method is compared with three competitive approaches, i.e. FaceTracer \cite{kumar2008facetracer}, PANDA-w \cite{zhang2013panda}, and PANDA-l \cite{zhang2013panda}. FaceTracer extracts HOG and color histograms in several important functional face regions and then trains SVM for attribute classification. We extract these functional regions referring to the ground truth landmark points.
PANDA-w and PANDA-l are based on PANDA \cite{zhang2013panda}, which was proposed recently for human attribute recognition by ensembling multiple CNNs, each of which extracts features from a well-aligned human part. These features are concatenated to train SVM for attribute recognition. It is straightforward to adapt this method to face attributes, since face parts can be well-aligned by landmark points. Here, we consider two settings. PANDA-w obtains the face parts by applying the state-of-the-art face detection \cite{li2013learning} and alignment \cite{sun2013deep} on wild images, while PANDA-l attains the face parts by using ground truth landmark points. For fair comparison, all the above methods are trained with the same data as ours.

\begin{figure*}[t]
  \centering
  \includegraphics[width=0.9\textwidth]{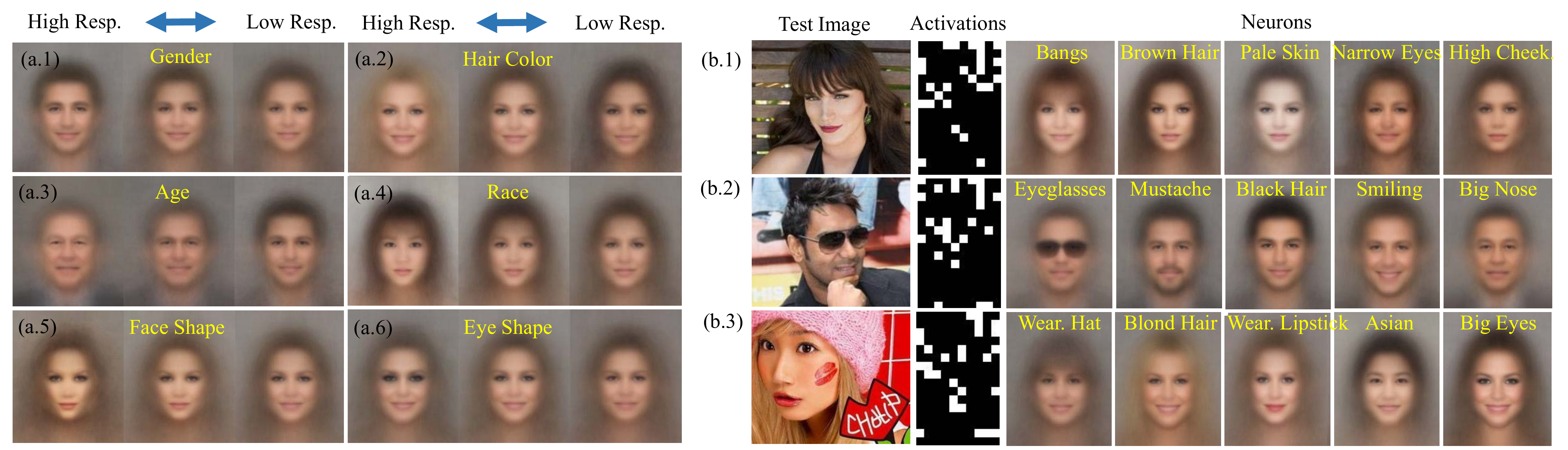}
  \caption{\footnotesize Visualization of neurons in ANet (a) after pre-training (b) after fine-tuning \textbf{(Best viewed in color)}}
  \vspace{-10pt}
  \label{fig:fnet}
\end{figure*}

\begin{figure}[t]
  \centering
  \includegraphics[width=0.45\textwidth]{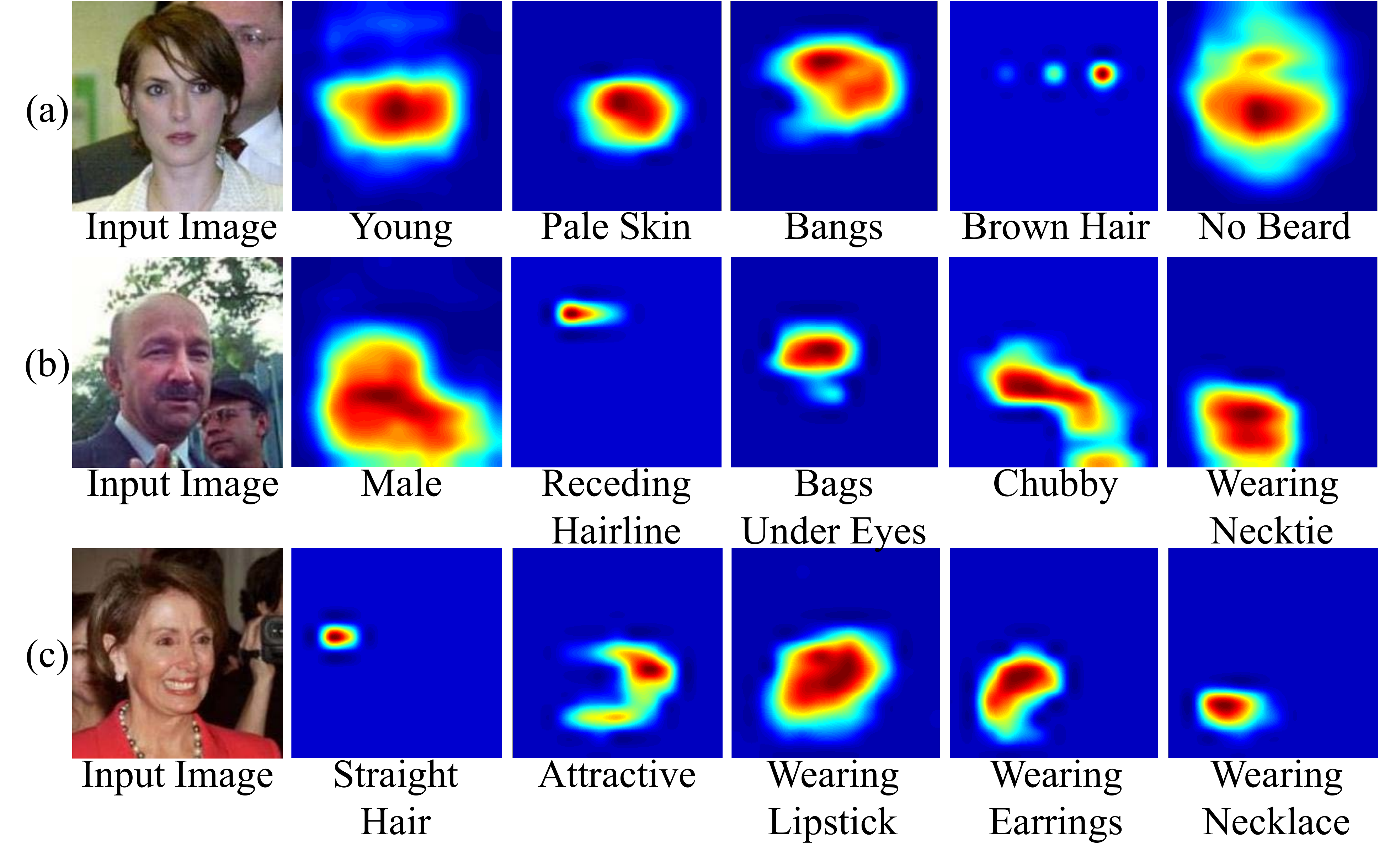}
  \caption{\footnotesize Attribute-specific regions discovery.}
  \label{fig:partsdiscovery}
\end{figure}

\subsection{Effectiveness of the Framework}

This section demonstrates the effectiveness of the framework. All experiments in this section are done on CelebA.

\textbf{$\bullet$ LNet}
%\subsubsection{LNet}
%
% We analyze LNet in two aspects.

\textbf{Performance Comparison}
We compare LNet with four state-of-the-art face detectors, including DPM \cite{mathias2014face}, ACF Multi-view \cite{yang2014aggregate}, SURF Cascade \cite{li2013learning}, and Face++ \cite{facepp}.
% on the task of face localization\footnote{Face localization assumes each image contains one face whose location is unknown. It is different from face detection which assumes an image could contain no face or multiple faces.}.
We evaluate them by using ROC curves when $IoU$\footnote{IoU indicates Intersection over Union.}$\geq$$0.5$.
%
% As plotted in Fig.\ref{fig:lnet}(a), when $FPPI = 0.1$, overall recalls of these five methods are $96\%$, $85\%$, $87\%$, $74\%$, and $85\%$ respectively. LNet outperforms them by 11, 9, 22, and 11 percent respectively.
%
As plotted in Fig.\ref{fig:lnet}(a), when $FPPI = 0.01$, the true positive rates of Face++ and LNet are $85\%$ and $93\%$; when $FPPI = 0.1$, our method outperforms the other three methods by 11, 9 and 22 percent respectively.
%
% Note that face localization for attribute prediction is different from general face detection.
%the standard metric of $IoU$\footnote{IoU indicates Intersection over Union.}$\geq$$0.5$ is employed.
%First, we evaluate how well the ``dominant'' face can be localized for attribute prediction when $FPPI$ is extremely small, \ie $FPPI = 0.01$.
%LNet outperforms the above methods by 14, 22, 35, and 7 percent, respectively,
%
%Furthermore, when $FPPI=0.1$,
We also investigate how these methods perform with respect to overlap ratio ($IoU$), following \cite{zitnick2014edge, mathias2014face}.
% This measure was used in \cite{zitnick2014edge, mathias2014face}.
Fig.\ref{fig:lnet}(c) shows that LNet generally provides more accurate face localization, leading to good performance in the subsequent attribute prediction.

\begin{figure}[t]
  \centering
  \includegraphics[width=0.45\textwidth]{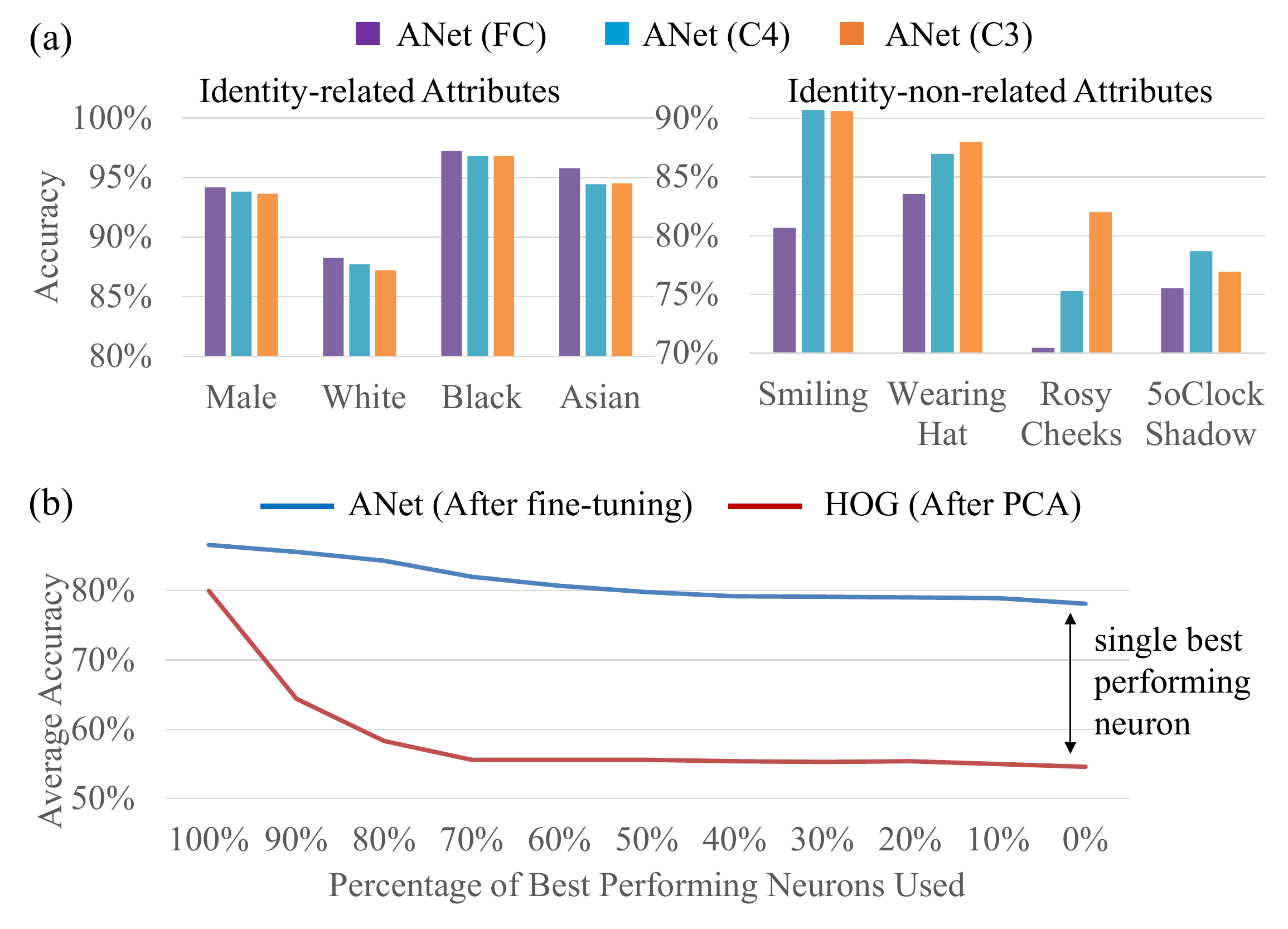}
  \caption{\footnotesize (a) Layer-wise comparison of ANet after pre-training (b) Best performing neurons analysis of ANet after fine-tuning. Best performing neurons are different for different attributes. The proposed accuracies are averaged over attributes which select their own subsets of best performing neurons.}
  \vspace{-10pt}
  \label{fig:layerwise}
\end{figure}

\begin{figure*}[t]
  \centering
  \includegraphics[width=\textwidth]{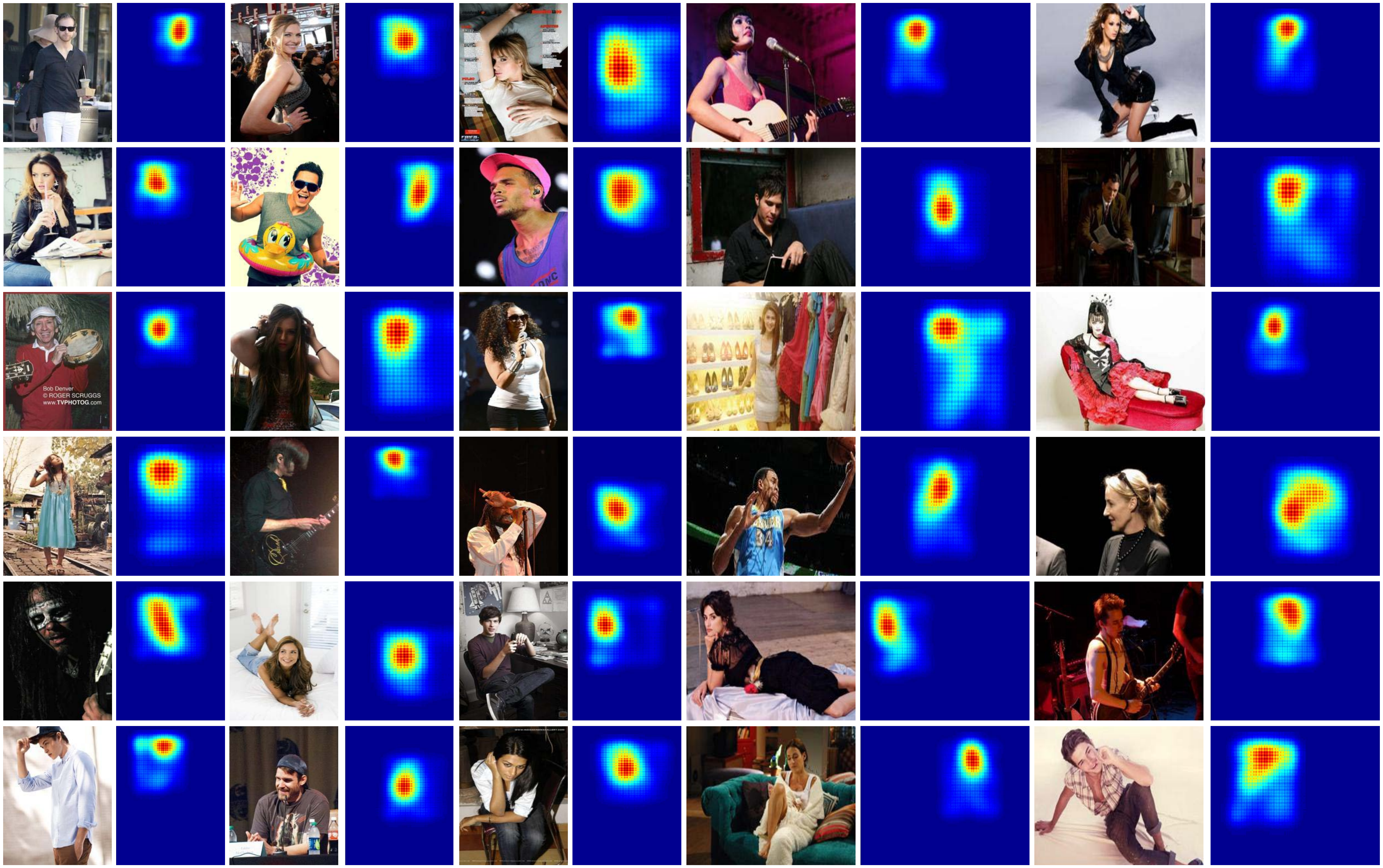}
  \caption{\footnotesize More results of LNet averaged response maps. \textbf{(Best viewed in color)}}
  \label{fig:localizationmap_more}
\end{figure*}

\textbf{Further Analysis}
LNet significantly outperforms LNet (without pre-training) by 74 percent when the overlap ratio equals to $0.5$, which validates the effectiveness of pre-training, as shown in Fig.\ref{fig:lnet}(c).
% pre-training improves face localization, as shown in Fig.\ref{fig:lnet}(c).
We then explore the influence of the number of attributes on localization.
Fig.\ref{fig:lnet}(d) illustrates rich attribute information facilitates face localization.
To examine the generalization ability of LNet, we collect another $3,876$ face images for testing, namely MobileFaces, which comes from a different source\footnote{MobileFaces was collected by normal users with mobile phones, while CelebA and LFWA collected face images of celebrities taken by professional photographers.} and has a different distribution from CelebA.
Several examples of MobileFaces are shown in Fig.\ref{fig:localizationmap}(b) and the corresponding ROC curves are plotted in Fig.\ref{fig:lnet}(b).
We observe that LNet constantly performs better and still gains 7 percent improvement ($FPPI=0.1$) compared with other face detectors.
% on this extended test set.
% Despite some failure cases due to extreme poses, large occlusions and low resolutions, LNet accurately localize faces in the wild
Despite some failure cases due to extreme poses and large occlusions, LNet accurately localize faces in the wild
% \footnote{Please refer to \textbf{supplementary materials} for more results.}
as demonstrated in Fig.\ref{fig:localizationmap}.
% Please refer to \textbf{supplementary materials} for more results.
%
More results of LNet under different circumstances (lighting, pose, occlusion, image resolution, background clutter \etc) are shown in Fig.\ref{fig:localizationmap_more}.

\textbf{Attribute-specific Regions Discovery}
Different attribute captures information from different region of face. We show that LNet automatically learns to discover these regions.
Given an attribute, by converting fully connected layers of LNet into fully convolutional layers following \cite{long2014fully}, we can locate important region of this attribute.
Fig.\ref{fig:partsdiscovery} shows some examples. The important regions of some attributes are locally distributed, such as `Bags Under Eyes', `Straight Hair' and `Wearing Necklace', but some are globally distributed, such as `Young', `Male' and `Attractive'.

\begin{figure}[t]
  \centering
  \includegraphics[width=0.45\textwidth]{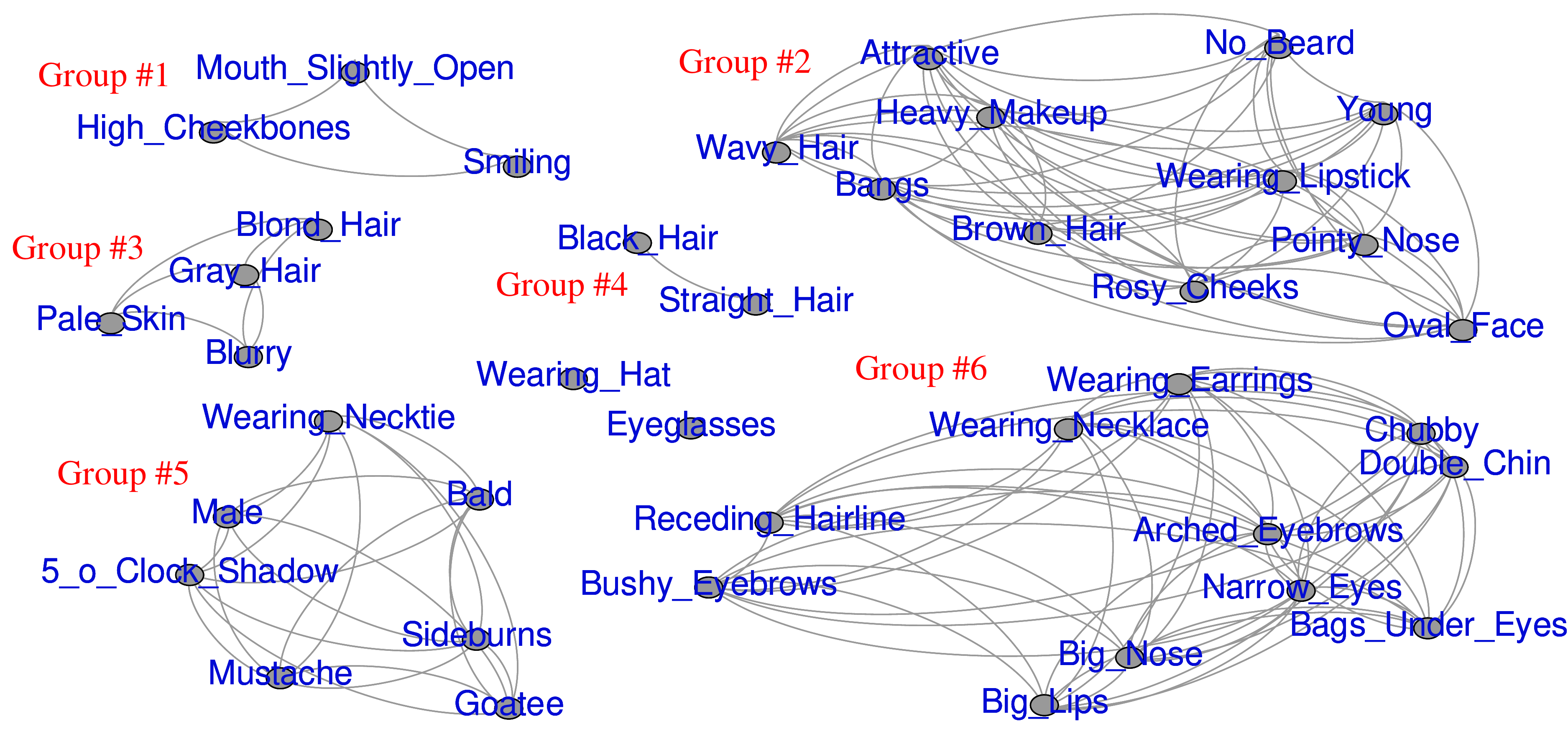}
  \caption{\footnotesize Automatic attributes grouping.}
  \label{fig:attributesgraph}
\end{figure}

\begin{table*}
\scriptsize
\begin{center}
\begin{tabular}{c|l|c|c|c|c|c|c|c|c|c|c|c|c|c|c|c|c|c|c|c|c|c|}
 & & \rotatebox{90}{5 Shadow} & \rotatebox{90}{Arch. Eyebrows} & \rotatebox{90}{Attractive} & \rotatebox{90}{Bags Un. Eyes} & \rotatebox{90}{Bald} & \rotatebox{90}{Bangs} & \rotatebox{90}{Big Lips} & \rotatebox{90}{Big Nose} & \rotatebox{90}{Black Hair} & \rotatebox{90}{Blond Hair} & \rotatebox{90}{Blurry} & \rotatebox{90}{Brown Hair} & \rotatebox{90}{Bushy Eyebrows} & \rotatebox{90}{Chubby} & \rotatebox{90}{Double Chin} & \rotatebox{90}{Eyeglasses} & \rotatebox{90}{Goatee} & \rotatebox{90}{Gray Hair} & \rotatebox{90}{Heavy Makeup} & \rotatebox{90}{H. Cheekbones} & \rotatebox{90}{Male} \\
\hline
\multirow{6}{*}{\rotatebox{360}{CelebA}} & FaceTracer \cite{kumar2008facetracer} & 85 & 76 & 78 & 76 & 89 & 88 & 64 & 74 & 70 & 80 & 81 & 60 & 80 & 86 & 88 & 98 & 93 & 90 & 85 & 84 & 91 \\
 & PANDA-w \cite{zhang2013panda} & 82 & 73 & 77 & 71 & 92 & 89 & 61 & 70 & 74 & 81 & 77 & 69 & 76 & 82 & 85 & 94 & 86 & 88 & 84 & 80 & 93 \\
 & PANDA-l \cite{zhang2013panda} & 88 & 78 & \bf{81} & \bf{79} & 96 & 92 & 67 & 75 & 85 & 93 & \bf{86} & 77 & 86 & 86 & 88 & 98 & 93 & 94 & \bf{90} & 86 & 97 \\
 & \cite{li2013learning}+ANet & 86 & 75 & 79 & 77 & 92 & 94 & 63 & 74 & 77 & 86 & 83 & 74 & 80 & 86 & 90 & 96 & 92 & 93 & 87 & 85 & 95 \\
 & LNets+ANet(w/o) & 88 & 74 & 77 & 73 & 95 & 92 & 66 & 75 & 84 & 91 & 80 & 78 & 85 & 86 & 88 & 96 & 92 & 93 & 85 & 84 & 94 \\
 & LNets+ANet & \bf{91} & \bf{79} & \bf{81} & \bf{79} & \bf{98} & \bf{95} & \bf{68} & \bf{78} & \bf{88} & \bf{95} & 84 & \bf{80} & \bf{90} & \bf{91} & \bf{92} & \bf{99} & \bf{95} & \bf{97} & \bf{90} & \bf{87} & \bf{98} \\
\hline\hline
\multirow{6}{*}{\rotatebox{360}{LFWA}} & FaceTracer \cite{kumar2008facetracer} & 70 & 67 & 71 & 65 & 77 & 72 & 68 & 73 & 76 & 88 & 73 & 62 & 67 & 67 & 70 & 90 & 69 & 78 & 88 & 77 & 84 \\
 & PANDA-w \cite{zhang2013panda} & 64 & 63 & 70 & 63 & 82 & 79 & 64 & 71 & 78 & 87 & 70 & 65 & 63 & 65 & 64 & 84 & 65 & 77 & 86 & 75 & 86 \\
 & PANDA-l \cite{zhang2013panda} & \bf{84} & 79 & 81 & 80 & 84 & 84 & 73 & 79 & 87 & 94 & 74 & 74 & 79 & 69 & 75 & 89 & 75 & 81 & 93 & 86 & 92 \\
 & \cite{li2013learning}+ANet & 78 & 66 & 75 & 72 & 86 & 84 & 70 & 73 & 82 & 90 & \bf{75} & 71 & 69 & 68 & 70 & 88 & 68 & 82 & 89 & 79 & 91 \\
 & LNets+ANet(w/o) & 81 & 78 & 80 & 79 & 83 & 84 & 72 & 76 & 86 & 94 & 70 & 73 & 79 & 70 & 74 & 92 & 75 & 81 & 91 & 83 & 91 \\
 & LNets+ANet & \bf{84} & \bf{82} & \bf{83} & \bf{83} & \bf{88} & \bf{88} & \bf{75} & \bf{81} & \bf{90} & \bf{97} & 74 & \bf{77} & \bf{82} & \bf{73} & \bf{78} & \bf{95} & \bf{78} & \bf{84} & \bf{95} & \bf{88} & \bf{94} \\
\hline\hline
 & & \rotatebox{90}{Mouth S. O.} & \rotatebox{90}{Mustache} & \rotatebox{90}{Narrow Eyes} & \rotatebox{90}{No Beard} & \rotatebox{90}{Oval Face} & \rotatebox{90}{Pale Skin} & \rotatebox{90}{Pointy Nose} & \rotatebox{90}{Reced. Hairline} & \rotatebox{90}{Rosy Cheeks} & \rotatebox{90}{Sideburns} & \rotatebox{90}{Smiling} & \rotatebox{90}{Straight Hair} & \rotatebox{90}{Wavy Hair} & \rotatebox{90}{Wear. Earrings} & \rotatebox{90}{Wear. Hat} & \rotatebox{90}{Wear. Lipstick} & \rotatebox{90}{Wear. Necklace} & \rotatebox{90}{Wear. Necktie} & \rotatebox{90}{Young} &  & \rotatebox{90}{\textbf{Average}} \\
\hline
\multirow{6}{*}{\rotatebox{360}{CelebA}} & FaceTracer \cite{kumar2008facetracer} & 87 & 91 & 82 & 90 & 64 & 83 & 68 & 76 & 84 & 94 & 89 & 63 & 73 & 73 & 89 & 89 & 68 & 86 & 80 & & 81 \\
 & PANDA-w \cite{zhang2013panda} & 82 & 83 & 79 & 87 & 62 & 84 & 65 & 82 & 81 & 90 & 89 & 67 & 76 & 72 & 91 & 88 & 67 & 88 & 77 & & 79 \\
 & PANDA-l \cite{zhang2013panda} & \bf{93} & 93 & \bf{84} & 93 & 65 & \bf{91} & 71 & 85 & 87 & 93 & \bf{92} & 69 & 77 & 78 & 96 & \bf{93} & 67 & 91 & 84 & & 85 \\
 & \cite{li2013learning}+ANet & 85 & 87 & 83 & 91 & 65 & 89 & 67 & 84 & 85 & 94 & \bf{92} & 70 & 79 & 77 & 93 & 91 & 70 & 90 & 81 & & 83 \\
 & LNets+ANet(w/o) & 86 & 91 & 77 & 92 & 63 & 87 & 70 & 85 & 87 & 91 & 88 & 69 & 75 & 78 & 96 & 90 & 68 & 86 & 83 & & 83 \\
 & LNets+ANet & 92 & \bf{95} & 81 & \bf{95} & \bf{66} & \bf{91} & \bf{72} & \bf{89} & \bf{90} & \bf{96} & \bf{92} & \bf{73} & \bf{80} & \bf{82} & \bf{99} & \bf{93} & \bf{71} & \bf{93} & \bf{87} & & \bf{87} \\
\hline\hline
\multirow{6}{*}{\rotatebox{360}{LFWA}} & FaceTracer \cite{kumar2008facetracer} & 77 & 83 & 73 & 69 & 66 & 70 & 74 & 63 & 70 & 71 & 78 & 67 & 62 & 88 & 75 & 87 & 81 & 71 & 80 & & 74 \\
 & PANDA-w \cite{zhang2013panda} & 74 & 77 & 68 & 63 & 64 & 64 & 68 & 61 & 64 & 68 & 77 & 68 & 63 & 85 & 78 & 83 & 79 & 70 & 76 & & 71 \\
 & PANDA-l \cite{zhang2013panda} & 78 & 87 & 73 & 75 & 72 & \bf{84} & 76 & 84 & 73 & 76 & 89 & 73 & 75 & 92 & 82 & 93 & 86 & \bf{79} & 82 & & 81 \\
 & \cite{li2013learning}+ANet & 76 & 79 & 74 & 69 & 66 & 68 & 72 & 70 & 71 & 72 & 82 & 72 & 65 & 87 & 82 & 86 & 81 & 72 & 79 & & 76 \\
 & LNets+ANet(w/o) & 78 & 87 & 77 & 75 & 71 & 81 & 76 & 81 & 72 & 72 & 88 & 71 & 73 & 90 & 84 & 92 & 83 & 76 & 82 & & 79 \\
 & LNets+ANet & \bf{82} & \bf{92} & \bf{81} & \bf{79} & \bf{74} & \bf{84} & \bf{80} & \bf{85} & \bf{78} & \bf{77} & \bf{91} & \bf{76} & \bf{76} & \bf{94} & \bf{88} & \bf{95} & \bf{88} & \bf{79} & \bf{86} & & \bf{84} \\
\hline\hline
\end{tabular}
\end{center}
\vspace{-5pt}
\caption{\footnotesize Performance comparison of attribute prediction. (Note that FaceTracer and PANDA-l attains the face parts by using ground truth landmark points.)}
\vspace{-10pt}
\label{tab:benchmarking}
\end{table*}

\textbf{$\bullet$ ANet}
%
% We analyze ANet in two aspects.

\textbf{Pre-training Discovers Semantic Concepts} 
We show that  pre-training of ANet can implicity discover semantic concepts related to face identity. Given a hidden neuron at the FC layer of ANet as shown in Fig.\ref{fig:pipeline}(c), we partition the face images into three groups, including the face images with high, medium, and low responses at this neuron. The face images of each group are then averaged to obtain the mean face.  We visualize these mean faces for several neurons in Fig.\ref{fig:fnet}(a). Interestingly, these mean face changes smoothly from high response to low response, following a high-level concept. Human can easily assign each neuron with a semantic concept it measures (\ie the text in yellow). For example, the neurons in (a.1) and (a.4) correspond to `gender' and `race', respectively. It reveals that the high-level hidden neurons of ANet can implicitly learn to discover semantic concepts, even though they are only optimized for face recognition using identity information and attribute labels are not used in pre-training.
We also observe that most of these concepts are intrinsic to face identity, such as the shape of facial components, gender, and race.

To better explain this phenomena, we compare the accuracy of attribute prediction using features at different layers of ANet right after pre-training. They are FC, C4, and C3. The forty attributes are roughly separated into two groups,
which are identity-related attributes, such as gender and race, and identity-non-related attributes, e.g. attributes of expressions, wearing hat and sunglasses.
We select some representative attributes for each group and plot the results in Fig.\ref{fig:layerwise}(a), which shows that the performance of FC outperforms C4 and C3 in the group of identity-related attributes, but they are relatively weaker when dealing with identity-non-related attributes. This is because the top layer FC learns identity features, which are insensitive to intra-personal face variations.

\textbf{Fine-tuning Expands Semantic Concepts}
%
%In the above, we have shown that pre-training of ANet essentially discovers semantic concepts related to identity.
Fig.\ref{fig:fnet} shows that after fine-tuning, ANet can expand these concepts to more attribute types.
Fig.\ref{fig:fnet}(b) visualizes the neurons in the FC layer, which are ranked by their responses in descending order with respect to several test images.
Human can assign semantic meaning to each of these neurons. We found that a large number of new concepts can be observed. Remarkably,
these neurons express diverse high-level meanings and cooperate to explain the test images.
The activations of all the neurons are visualized in Fig.\ref{fig:fnet}(b), and they are sparse.
In some sense, attributes presented in each test image are explained by a sparse linear combination of these concepts.
For instance, the first image is described as ``a lady with bangs, brown hair, pale skin, narrow eyes and high cheekbones'', which well matches human perception.

To validate this, we explore how the number of neurons influences attribute prediction accuracies.
Best performing neurons for each attribute are identified by sorting corresponding SVM weights.
Fig.\ref{fig:layerwise}(b) illusatrates that only $10\%$ of ANet best performing neurons are needed to achieve $90\%$ of the original performance of a particular attribute\footnote{Best performing neurons are different for different attributes.}. In contrast, HOG+PCA does not have the sparse nature and need more than $95\%$ features
Besides, the best single performing neuron of ANet outperforms that of HOG+PCA by $25$ percent in average prediction accuracy.

%\textbf{Attribute Grouping}
%%
%Here we show that the weight matrix at the FC layer of ANet can implicitly capture relations between attributes. Each column vector of the weight matrix can be viewed as a decision hyperplane to partition the negatives and positive samples of an attribute. By simply applying k-means to these vectors, the clusters show clear grouping patterns, which can be interpreted semantically.
%%
%As shown in Fig.\ref{fig:attributesgraph}, Group \#1, Group \#2 and Group \#5 demonstrate co-occurence relationship between attributes, \eg 'Black Hair' and 'Straigt Hair' have high correlation.
%Attributes in Group \#3 share similar color discriptors while attributes in Group \#6 correspond to certain texture and appearance traits.

\begin{table*}
\scriptsize
\begin{center}
\begin{tabular}{l|c|c|c|c|c|c|c|c|c|c|c|c|c|c|c|c|c|c|c|c|c|}
 & \rotatebox{90}{Gender} & \rotatebox{90}{Asian} & \rotatebox{90}{White} & \rotatebox{90}{Black} & \rotatebox{90}{Youth} & \rotatebox{90}{M. Aged} & \rotatebox{90}{Senior} & \rotatebox{90}{Black H.} & \rotatebox{90}{Blond H.} & \rotatebox{90}{Bald} & \rotatebox{90}{No Eye.} & \rotatebox{90}{Eye.} & \rotatebox{90}{Mustache} & \rotatebox{90}{R. Hair.} & \rotatebox{90}{B. Eye.} & \rotatebox{90}{A. Eye.} & \rotatebox{90}{B. Nose} & \rotatebox{90}{No Beard} & \rotatebox{90}{R. Jaw} & & \rotatebox{90}{\textbf{Average}} \\
\hline
 FaceTracer \cite{kumar2008facetracer} & 91 & \bf{87} & \bf{86} & 75 & 66 & 54 & 70 & 66 & 68 & 72 & 84 & 86 & 83 & 76 & 72 & 66 & 65 & \textbf{81} & 51 &  & 73 \\
 POOF \cite{berg2013poof} & 92 & 90 & 81 & \bf{90} & 71 & 60 & 80 & 67 & 75 & 67 & \bf{87} & \bf{90} & \bf{86} & 72 & 74 & 71 & 68 & 77 & 55 & & 76 \\
 LNets+ANet & \bf{94} & 85 & 83 & 87 & \bf{80} & \bf{77} & \bf{81} & \bf{86} & \bf{89} & \bf{84} & 85 & 84 & \bf{86} & \bf{83} & \bf{82} & \bf{75} & \bf{79} & 78 & \bf{81} & & \bf{83} \\
\hline\hline
\end{tabular}
\end{center}
\vspace{-5pt}
\caption{\footnotesize Performance comparison on extended attributes. (Performance are measured by the average of true positive rates and true negative rates.)}
\vspace{-10pt}
\label{tab:extended}
\end{table*}

\textbf{Automatic Attributes Grouping}
Here we show that the weight matrix at the FC layer of ANet can implicitly capture relations between attributes. Each column vector of the weight matrix can be viewed as a decision hyperplane to partition the negatives and positive samples of an attribute. By simply applying k-means to these vectors, the clusters show clear grouping patterns, which can be interpreted semantically.
As shown in Fig.\ref{fig:attributesgraph}, Group \#1, Group \#2 and Group \#4 demonstrate co-occurrence relationship between attributes, \eg `Attractive' and `Heavy Makeup' have high correlation.
Attributes in Group \#3 share similar color descriptors, while attributes in Group \#6 correspond to certain texture and appearance traits.

\begin{figure}[t]
  \centering
  \includegraphics[width=0.5\textwidth]{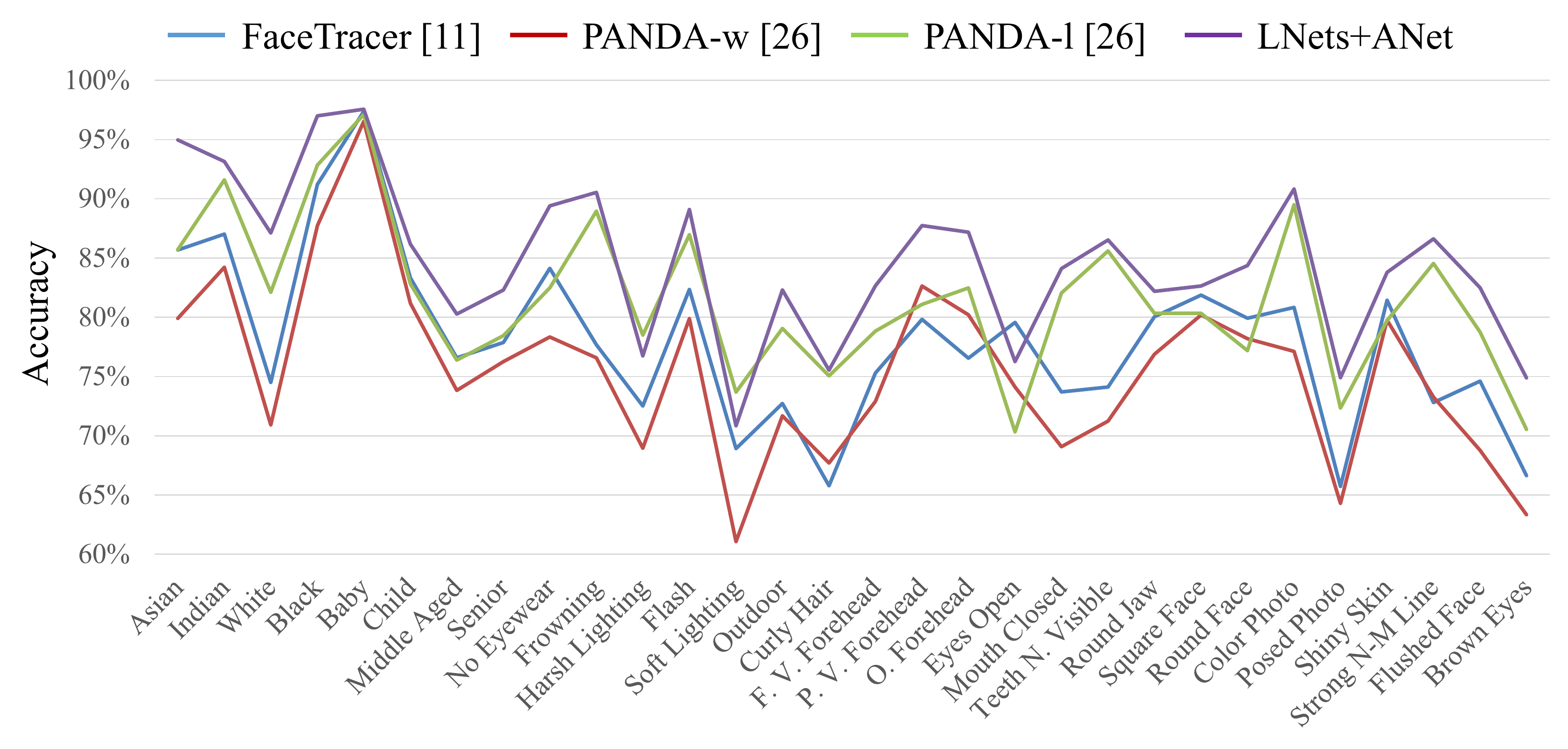}
  \caption{\footnotesize Performance comparison of FaceTracer \cite{kumar2008facetracer}, PANDA-w \cite{zhang2013panda}, PANDA-l \cite{zhang2013panda} and LNets+ANet on LFWA+.}
  \label{fig:benchmark_transfer}
\end{figure}

\begin{figure}[t]
  \centering
  \includegraphics[width=0.5\textwidth]{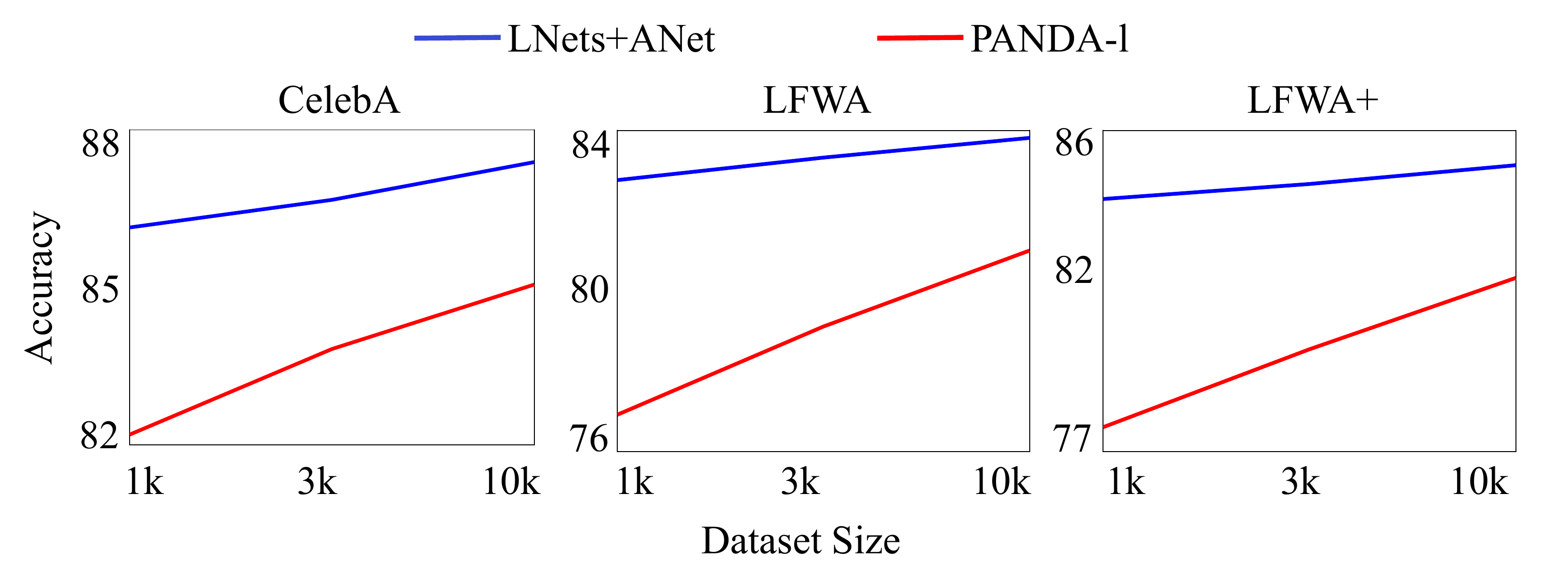}
  \caption{\footnotesize Performances of different training dataset sizes.}
  \label{fig:datasetsize}
\end{figure}

\subsection{Attribute Prediction}

\textbf{Performance Comparison}
The attribute prediction performance is reported in Table.\ref{tab:benchmarking}.
On CelebA, the prediction accuracies of FaceTracer \cite{kumar2008facetracer}, PANDA-w \cite{zhang2013panda}, PANDA-l \cite{zhang2013panda}, and our LNets+ANet are $81$, $79$, $85$, and $87$ percent respectively,
while the corresponding accuracies on LFWA are $74$, $71$, $81$, and $84$ percent.
Our method outperforms PANDA-w by nearly $10$ percent.
Remarkably, even when PANDA-l is equipped with groundtruth bounding boxes and landmark positions, our method still achieves $3$ percent gain.
The strength of our method is illustrated not only on global attributes, \eg ``Chubby'' and ``Young'', but also on fine-grained facial traits, \eg ``Mastache'' and ``Pointy Nose''.
We also report performance on $19$ extended attributes and compare our result with \cite{kumar2008facetracer} and \cite{berg2013poof}. 
%by directly transferring weights learned above.
The evaluation protocol is the same as \cite{berg2013poof}.
%Results of \cite{kumar2008facetracer} and \cite{berg2013poof} are taken from their papers.
In Table \ref{tab:extended}, LNets+ANet outperforms them by $10$ and $7$ percent respectively.

\textbf{Further Analysis}
When compared with \cite{li2013learning}+ANet, LNets accounts for nearly $6$ percentage improvement over using an off-the-shelf face detector \cite{li2013learning}.
We also experiment with the case of providing ANet with localized face region by LNets, but without pre-training, denoted as LNets+ANet(w/o).
The average accuracies have dropped $4$ and $5$ percent on CelebA and LFWA, which indicate pre-training with massive facial identities helps discover semantic concepts.

\textbf{Performance on LFWA+}
To further examine whether the proposed approach can be generalized to unseen attributes, we manually label $30$ more attributes for the testing images on LFWA and denote this extended dataset as LFWA+.
To test on these $30$ attributes, we directly transfer weights learned by deep models to extract features, and only re-train SVMs using one third of the images.
% These $30$ attributes do not appear in training.
LNets+ANet leads to $8$, $10$, and $3$ percent average gains over the other three approaches (FaceTracer, PANDA-w, and PANDA-l).
It demonstrates that our method learns discriminative face representations and has good generalization ability.

\textbf{Size of Training Dataset}
%
%In the practical usage, it is often expensive or infeasible to obtain enough training data for deep models.
We compare the attribute prediction accuracy of the proposed method with the accuracy of PANDA-l, regarding different sizes of training datasets.
% when the provided training dataset size changes.
Only the training data of ANet is changed in our method for fair comparison.
Fig.\ref{fig:datasetsize} demonstrates that LNets+ANet performs well when dataset size is small, but the performance of PANDA-l drops significantly.
%The pre-training process renders our model a good initialization point.
%Thus, our method is also suitable for other face analyzing tasks when the available data is not big enough.

\textbf{Time Complexity}
For a $300*300$ image, LNets takes $35$ms to localize face region while ANet takes $14$ms to extract features on GPU.
In contrast, a na{\"i}ve patch-by-patch scanning needs nearly $80$ ms to extract features.
Our framework has large potential in real-world applications.
%More detailed results are in \textbf{supplementary materials}.

%------------------------------------------------------------------------
\section{Conclusion}

This paper has proposed a novel deep learning framework for face attribute prediction in the wild.
With carefully designed pre-training strategies, our method is robust to background clutters and face variations.
We devise a new fast feed-forward algorithm for locally shared filters to save redundant computation, which enables evaluating image with arbitrary size in realtime. It allows taking images of arbitrary sizes as input without normalization.
We have also revealed multiple important facts about learning face representation, which shed a light on new directions of face localization and representation learning.

{\small
\bibliographystyle{ieee}
\bibliography{egbib}

\begin{thebibliography}{10}\itemsep=-1pt

\bibitem{facepp}
Face++.
\newblock \url{http://www.faceplusplus.com/}.

\bibitem{berg2013poof}
T.~Berg and P.~N. Belhumeur.
\newblock Poof: Part-based one-vs.-one features for fine-grained
  categorization, face verification, and attribute estimation.
\newblock In {\em CVPR}, pages 955--962, 2013.

\bibitem{bergamo2014self}
A.~Bergamo, L.~Bazzani, D.~Anguelov, and L.~Torresani.
\newblock Self-taught object localization with deep networks.
\newblock {\em arXiv preprint arXiv:1409.3964}, 2014.

\bibitem{bourdev2011describing}
L.~Bourdev, S.~Maji, and J.~Malik.
\newblock Describing people: A poselet-based approach to attribute
  classification.
\newblock In {\em ICCV}, pages 1543--1550, 2011.

\bibitem{chung2012deep}
J.~Chung, D.~Lee, Y.~Seo, and C.~D. Yoo.
\newblock Deep attribute networks.
\newblock In {\em NIPS Workshop on Deep Learning and Unsupervised Feature
  Learning}, volume~3, 2012.

\bibitem{deng2009imagenet}
J.~Deng, W.~Dong, R.~Socher, L.-J. Li, K.~Li, and L.~Fei-Fei.
\newblock Imagenet: A large-scale hierarchical image database.
\newblock In {\em CVPR}, pages 248--255, 2009.

\bibitem{donahue2013decaf}
J.~Donahue, Y.~Jia, O.~Vinyals, J.~Hoffman, N.~Zhang, E.~Tzeng, and T.~Darrell.
\newblock Decaf: A deep convolutional activation feature for generic visual
  recognition.
\newblock {\em arXiv preprint arXiv:1310.1531}, 2013.

\bibitem{REF08a}
R.-E. Fan, K.-W. Chang, C.-J. Hsieh, X.-R. Wang, and C.-J. Lin.
\newblock {Liblinear}: A library for large linear classification.
\newblock {\em JMLR}, 9:1871--1874, 2008.

\bibitem{farhadi2009describing}
A.~Farhadi, I.~Endres, D.~Hoiem, and D.~Forsyth.
\newblock Describing objects by their attributes.
\newblock In {\em CVPR}, pages 1778--1785, 2009.

\bibitem{hadsell2006dimensionality}
R.~Hadsell, S.~Chopra, and Y.~LeCun.
\newblock Dimensionality reduction by learning an invariant mapping.
\newblock In {\em CVPR}, volume~2, pages 1735--1742, 2006.

\bibitem{he2014spatial}
K.~He, X.~Zhang, S.~Ren, and J.~Sun.
\newblock Spatial pyramid pooling in deep convolutional networks for visual
  recognition.
\newblock In {\em ECCV}, pages 346--361. 2014.

\bibitem{LFWTech}
G.~B. Huang, M.~Ramesh, T.~Berg, and E.~Learned-Miller.
\newblock Labeled faces in the wild: A database for studying face recognition
  in unconstrained environments.
\newblock Technical Report 07-49, University of Massachusetts, Amherst, October
  2007.

\bibitem{krizhevsky2012imagenet}
A.~Krizhevsky, I.~Sutskever, and G.~E. Hinton.
\newblock Imagenet classification with deep convolutional neural networks.
\newblock In {\em NIPS}, pages 1097--1105, 2012.

\bibitem{kumar2008facetracer}
N.~Kumar, P.~Belhumeur, and S.~Nayar.
\newblock Facetracer: A search engine for large collections of images with
  faces.
\newblock In {\em ECCV}, pages 340--353. 2008.

\bibitem{kumar2009attribute}
N.~Kumar, A.~C. Berg, P.~N. Belhumeur, and S.~K. Nayar.
\newblock Attribute and simile classifiers for face verification.
\newblock In {\em ICCV}, pages 365--372, 2009.

\bibitem{le1990handwritten}
Y.~LeCun, B.~Boser, J.~S. Denker, D.~Henderson, R.~E. Howard, W.~Hubbard, and
  L.~D. Jackel.
\newblock Handwritten digit recognition with a back-propagation network.
\newblock In {\em NIPS}, 1990.

\bibitem{li2013learning}
J.~Li and Y.~Zhang.
\newblock Learning surf cascade for fast and accurate object detection.
\newblock In {\em CVPR}, pages 3468--3475, 2013.

\bibitem{long2014fully}
J.~Long, E.~Shelhamer, and T.~Darrell.
\newblock Fully convolutional networks for semantic segmentation.
\newblock In {\em CVPR}, 2015.

\bibitem{luo2013deep}
P.~Luo, X.~Wang, and X.~Tang.
\newblock A deep sum-product architecture for robust facial attributes
  analysis.
\newblock In {\em ICCV}, pages 2864--2871, 2013.

\bibitem{manyam2011two}
O.~K. Manyam, N.~Kumar, P.~Belhumeur, and D.~Kriegman.
\newblock Two faces are better than one: Face recognition in group photographs.
\newblock In {\em IJCB}, pages 1--8, 2011.

\bibitem{mathias2014face}
M.~Mathias, R.~Benenson, M.~Pedersoli, and L.~Van~Gool.
\newblock Face detection without bells and whistles.
\newblock In {\em ECCV}, pages 720--735. 2014.

\bibitem{oquab2015object}
M.~Oquab, L.~Bottou, I.~Laptev, and J.~Sivic.
\newblock Is object localization for free?--weakly-supervised learning with
  convolutional neural networks.
\newblock In {\em CVPR}, pages 685--694, 2015.

\bibitem{razavian2014cnn}
A.~S. Razavian, H.~Azizpour, J.~Sullivan, and S.~Carlsson.
\newblock Cnn features off-the-shelf: an astounding baseline for recognition.
\newblock {\em arXiv preprint arXiv:1403.6382}, 2014.

\bibitem{rodriguez2014clustering}
A.~Rodriguez and A.~Laio.
\newblock Clustering by fast search and find of density peaks.
\newblock {\em Science}, 344(6191):1492--1496, 2014.

\bibitem{song2014exploiting}
F.~Song, X.~Tan, and S.~Chen.
\newblock Exploiting relationship between attributes for improved face
  verification.
\newblock {\em CVIU}, 122:143--154, 2014.

\bibitem{sun2013deep}
Y.~Sun, X.~Wang, and X.~Tang.
\newblock Deep convolutional network cascade for facial point detection.
\newblock In {\em CVPR}, pages 3476--3483, 2013.

\bibitem{sun2014deep}
Y.~Sun, X.~Wang, and X.~Tang.
\newblock Deep learning face representation by joint
  identification-verification.
\newblock In {\em NIPS}, 2014.

\bibitem{taigman2014deepface}
Y.~Taigman, M.~Yang, M.~Ranzato, and L.~Wolf.
\newblock Deepface: Closing the gap to human-level performance in face
  verification.
\newblock In {\em CVPR}, pages 1701--1708, 2014.

\bibitem{xiao2010sun}
J.~Xiao, J.~Hays, K.~A. Ehinger, A.~Oliva, and A.~Torralba.
\newblock Sun database: Large-scale scene recognition from abbey to zoo.
\newblock In {\em CVPR}, pages 3485--3492, 2010.

\bibitem{yang2014aggregate}
B.~Yang, J.~Yan, Z.~Lei, and S.~Z. Li.
\newblock Aggregate channel features for multi-view face detection.
\newblock In {\em IJCB}, pages 1--8, 2014.

\bibitem{zhang2014part}
N.~Zhang, J.~Donahue, R.~Girshick, and T.~Darrell.
\newblock Part-based r-cnns for fine-grained category detection.
\newblock In {\em ECCV}, pages 834--849. 2014.

\bibitem{zhang2013panda}
N.~Zhang, M.~Paluri, M.~Ranzato, T.~Darrell, and L.~Bourdev.
\newblock Panda: Pose aligned networks for deep attribute modeling.
\newblock In {\em CVPR}, 2014.

\bibitem{zhou2014object}
B.~Zhou, A.~Khosla, A.~Lapedriza, A.~Oliva, and A.~Torralba.
\newblock Object detectors emerge in deep scene cnns.
\newblock In {\em ICLR}, 2015.

\bibitem{zitnick2014edge}
C.~L. Zitnick and P.~Doll{\'a}r.
\newblock Edge boxes: Locating object proposals from edges.
\newblock In {\em ECCV}, pages 391--405. 2014.

\end{thebibliography}
}

\end{document}